%% file: main.tex
\documentclass{article}

\usepackage[a4paper,margin=1in]{geometry}

\usepackage{amsmath}
\usepackage{amssymb}
\usepackage{mathrsfs}

\usepackage{amsthm}

\usepackage{authblk}
\usepackage{parskip}
\usepackage{graphicx}    
\usepackage{placeins}    
\usepackage[export]{adjustbox}  
\usepackage{rotating}
\usepackage{color}
\usepackage{epsfig}
\usepackage{verbatim}
\usepackage{natbib}      
\usepackage[printonlyused]{acronym} 
\usepackage{setspace}    

\usepackage[capitalise]{cleveref}
\usepackage{bm}
\usepackage{booktabs}
\usepackage{multirow}
\usepackage[capposition=top]{floatrow}
\usepackage{afterpage}
\usepackage{relsize}
\usepackage{subcaption}
\usepackage[T1]{fontenc}

\usepackage{changepage} 
\usepackage{enumerate}

\input{defs}

\newtheorem{theorem}{Theorem}

\newtheorem{corollary}[theorem]{Corollary}
\theoremstyle{definition}
\newtheorem{definition}{Definition}
\theoremstyle{definition}
\newtheorem{assumption}{Assumption}

\title{Image Response Regression via Deep Neural Networks}

\author[1,2]{Daiwei Zhang}
\author[3]{Lexin Li}
\author[1]{Chandra Sripada}
\author[1,*]{Jian Kang}

\affil[1]{University of Michigan, Ann Arbor}
\affil[2]{University of Pennsylvania}
\affil[3]{University of California, Berkeley}
\affil[*]{\footnotesize Corresponding author. Email: jiankang@umich.edu}

\begin{document}

\maketitle

\input{abstract.tex}
\input{introduction.tex}
\input{model.tex}
\input{theory.tex}
\input{simulation.tex}
\input{realdata.tex}

\input{discussion.tex}

\newpage
\bibliographystyle{apalike}
\bibliography{refs}

\end{document}

%% file: defs.tex
\def\bfa{\bm{a}}
\def\bfb{\bm{b}}

\def\bfs{\bm{s}}
\def\bfu{\bm{u}}

\def\bfx{\bm{x}}
\def\bfy{\bm{y}}

\def\bfI{\bm{I}}

\def\bfW{\bm{W}}
\def\bfX{\bm{X}}

\def\bfalpha{\bm{\alpha}}
\def\bfbeta{\bm{\beta}}

\def\bfeta{\bm{\eta}}
\def\bftheta{\bm{\theta}}

\def\bfphi{\bm{\phi}}

\def\bbI{\mathbb I}
\def\bbN{\mathbb N}
\def\bbR{\mathbb R}

\def\calG{\mathcal G}
\def\calN{\mathcal N}
\def\calO{\mathcal O}
\def\calS{\mathcal S}

\def\E{\operatorname{E}}

\def\Cov{\operatorname{Cov}}

\def\sign{\operatorname{sign}}

\def\aleph{\mathscr{N}}

%% file: abstract.tex
\begin{abstract}
  Delineating the associations between images and a vector of covariates is of central interest in medical imaging studies. To tackle this problem of image response regression, we propose a novel nonparametric approach in the framework of spatially varying coefficient models, where the spatially varying functions are estimated through deep neural networks. Compared to existing solutions, the proposed method explicitly accounts for spatial smoothness and subject heterogeneity, has straightforward interpretations, and is highly flexible and accurate in capturing complex association patterns. A key idea in our approach is to treat the image voxels as the effective samples, which not only alleviates the limited sample size issue that haunts the majority of medical imaging studies, but also leads to more robust and reproducible results. Focusing on a broad family of piecewise smooth functions, we establish the estimation and selection consistency, and derive the asymptotic error bounds. We demonstrate the efficacy of the method through intensive simulations, and further illustrate its advantages with analyses of two functional magnetic resonance imaging datasets. 
\end{abstract}

%% file: introduction.tex
\section{Introduction}

A central question in medical imaging studies is to delineate the associations between images and a set of covariates, such as demographics and clinical features. For instance, anatomical magnetic resonance imaging (MRI) studies can be conducted to locate brain regions exhibiting structural differences between normal controls and patients with neurological disorders, after accounting for demographic covariates. Another example is functional magnetic resonance imaging (fMRI) studies for identifying brain locations that demonstrate differential activation patterns under different tasks. These examples can be collectively formulated as a regression problem where the image scans, usually measured over a two- or three-dimensional spatial space, are treated as the response variable, while the demographic and clinical features are treated as the predictors. We refer to this type of analysis as the image response regression. 

Image response regression involves multiple challenges. First, the image dimensions are ultrahigh. For instance, a $64 \times 64 \times 64$ MRI image records data over $64^3 = 262,144$ voxels. Meanwhile, complex spatial correlations exist among imaging voxels and regions. Second, the association patterns between images and covariates can be highly complex, as imaging signals are often found in contiguous, sharp-edged regions that are sparsely distributed throughout the spatial volume \citep{chumbley2009false}, which is exacerbated by a high noise level, all of which leads to a low signal-to-noise ratio.  Third, while a typical imaging study collects data from multiple subjects, the number of subjects is usually limited (mostly in the order of tens or hundreds) compared to the ultrahigh dimensionality of the image, with subject-specific heterogeneity being the rule rather than the exception. To address those challenges, there have been several lines of research proposed for image response regression. 

The first and perhaps the most common solution is the mass univariate analysis \citep[MUA]{friston2003statistical}, which fits a generalized linear model to associate the image measure at each voxel with the covariates. A major limitation of MUA is that the model is fitted one voxel at a time, without accounting for spatial correlations, which often leads to low detection power. To partially overcome this limitation, there have been proposals of pre-smoothing \citep{friston2003statistical}, or adaptive smoothing \citep{qiu2007jump, yue2010adaptive} to incorporate neighboring voxel information. Nevertheless, it remains challenging to capture complex associations between images and covariates, as well as subject heterogeneity. 

The second line of research treats the image as a tensor, i.e., a multi-dimensional array, and then fits a tensor response regression model under different low-rank and sparsity structures \citep{rabusseau2016, li2017parsimonious, sun2017store, raskutti2019convex, chen2019non}. These solutions implicitly accounts for spatial smoothness, but usually not subject heterogeneity. Moreover, due to the limited sample size, large noise, and computational constraints, the response images are often downsized to a smaller dimension to reach a compromise between model accuracy and feasibility. 

A third strategy is to utilize a spatially varying coefficient model, where the response images and the regression coefficients are modeled as the realizations of some spatially varying functions evaluated on discrete grid points, i.e., voxels, in a spatial domain. This family of solutions systematically incorporate spatial smoothness, jump discontinuities, and subject-specific heterogeneity, and have been shown to be highly effective in characterizing the image-covariate associations. Notably, \citet{zhu2014spatially} developed the spatially varying coefficient model for neuroimaging data with jump discontinuities. They employed local linear regression and functional principal components to estimate the coefficient functions, and developed an asymptotic Wald test to identify significant association regions. \citet{chen2016local} extended the model of \citet{zhu2014spatially}, by coupling with an $L_1$ type penalty for sparsity, and a total variation type penalty for spatial contiguity. \citet{LiZhu2017} introduced a single-index structure into the model, and iteratively estimated varying coefficient functions, link functions, index parameter vectors, and the covariance function of individual functions. Alternatively, \citet{li2020sparse} and \citet{yu2020multivariate} considered a version of the spatially varying coefficient model where the image is located in a two-dimensional space, but the covariates can be ultrahigh-dimensional. They employed bivariate splines over triangulation to approximate the coefficient functions, and added the adaptive $L_1$ penalty for simultaneous sparse learning and model structure identification. From a Bayesian perspective, \citet{shi2015thresholded} used thresholded multiscale Gaussian processes, while \citet{bussas2017varying} used Gaussian processes with isotropic priors, to model the spatially varying functions. 

In recent years, a family of machine learning methods based upon deep neural networks (DNN) have enjoyed enormous success in a wide range of applications, among which medical imaging analysis is a particularly remarkable example \citep{lecun2015deep, goodfellow2016deep}. Such success can be attributed to the expressive power of DNN in approximating and learning highly complex nonparametric models \citep{fan2019selective}. In fact, any continuous function on a compact set can be approximated by a single-hidden layer neural network with a sufficiently large number of nodes to an arbitrary degree of accuracy \citep{barron1994approximation}. Meanwhile, in terms of the required total number of nodes, a deeper neural network, i.e., one with more hidden layers, has been shown to be more efficient in approximating functions than a shallower one \citep{telgarsky2016benefits, eldan2016power}. Recently, studies have investigated the convergence rate of multi-layer neural networks. Notably, \citet{barron1994approximation, mccaffrey1994convergence} obtained the convergence rate for a single-layer neural network. \citet{kohler2017nonparametric} established the minimax convergence rate for a two-layer network. \citet{bauer2019deep} extended this result to a multi-layer network with a smooth activation function, whereas \citet{schmidt2020nonparametric} proved a similar result for a rectified linear unit (ReLU) activation function. Other studies have investigated variable selection in DNN. Notably, \citet{feng2017sparse} imposed a group $L_1$ penalty on the weights of the first layer, while \citet{chen2020nonlinear} introduced an additional selection layer to select relevant input variables.

Despite the wide success of DNN in medical imaging analysis, numerous challenges remain, one of which is the limited sample size. DNN training typically requires a fairly large sample size, while most medical imaging studies only enroll a small number of subjects. As a result, a successful application of DNN in medical imaging always requires intensive and arduous tuning of the architecture and parameters of the neural network, and is often difficult to reproduce. Aggregating multiple imaging centers to reach a large sample size is a viable solution, but at the cost of introducing center-specific variability. In addition, how to interpret the DNN model and how to handle data heterogeneity in DNN remain partially addressed only. 

In this article, we propose a novel image response regression approach that integrates spatially varying coefficient models with deep neural networks. That is, we adopt the spatially varying coefficient model framework similar to that of \citet{zhu2014spatially, li2020sparse}, but we estimate the functions of main effect, individual deviation, and error variance all through multi-layer neural networks. Our approach enjoys several advantages. Compared to the massive univariate analysis or tensor response regressions, our method explicitly accounts for both spatial smoothness and subject heterogeneity. We achieve the former by focusing our candidate functions in a piecewise smooth function class, and the latter by including and estimating an individual deviation function specific for each subject. Compared to the spatially varying coefficient models based on local linear regression or bivariate splines, our DNN-based model is more flexible and able to capture more complex association patterns, thanks to the approximation capability of DNNs. Moreover, we treat each subject-specific deviation function as a fixed target, rather than a random effect term, and estimate it directly, which allows a more direct characterization of the subject-specific heterogeneity. This treatment differentiates our proposal from alternative varying coefficient model-based solutions such as \citet{zhu2014spatially, li2020sparse}. Compared to the vast DNN-based medical imaging analysis literature, our model is straightforward to interpret, as we use DNNs to approximate the main effect, individual deviation, and error variance functions, while their interpretations remain the same as in a classical varying coefficient model. To further improve the interpretability of the main effect function, we impose sparsity on the entire function, but \emph{not} on the parameters of the fitted neural networks, bypassing the interpretation and selection of those individual parameters, which can be difficult. In addition, a key innovation of our proposal is that, when fitting the neural networks to those functions, we treat each spatial coordinate as a sample observation. Consequently, the effective sample size is no longer the number of subjects, but rather the total number of voxels, which is typically much larger. This strategy leads to a more accurate DNN fit to the data and a more robust and reproducible performance with repect to the architecture of the DNNs, which is confirm by our empirical studies.

Our proposal makes several useful contributions. We extend the family of spatially varying coefficient models to address an important class of scientific problems of image response regression. Our model is able to capture a wide variety of spatial correlation patterns, including both smooth transitions, as well as jump discontinuities. We also extend the theory of nonparametric regression modeling. We derive the asymptotic convergence rate of the estimated functions, and establish the selection consistency of the sparse function, as either the number of voxels or the number of subjects diverges to infinity. We derive the asymptotic error bound, not only for globally smooth functions, but also for a broader family of piecewise smooth generalized hierarchical interaction functions, and show that the bound is comparable to the minimax bound in the traditional nonparametric regressions. Finally, our proposal exemplifies how to utilize state-of-the-art deep learning tools to help address a classical statistical problem, which enjoys both modeling flexibility and theoretical guarantees.

The rest of the article is organized as follows. We present the model and the DNN-based estimation method in \Cref{sec:model-estimation}, followed by a study of their theoretical properties in \Cref{sec:theory}. We evaluate the empirical performance of our approach compared with alternative solutions in \Cref{sec:simulations}, and apply them to two brain imaging datasets in \Cref{sec:realdata}. \Cref{sec:discussion} concludes the article with a discussion. All technical proofs are relegated to the Supplementary Materials.

%% file: model.tex
\section{Model and Estimation}
\label{sec:model-estimation}

In this section, we first present the image response regression through the spatially varying coefficient model. We next propose to estimate the key functions in the model using deep neural networks, and derive a multi-step estimation procedure. 

\subsection{Image response regression with spatially varying coefficients}
\label{sec:model}

Suppose the imaging data are collected from a $D$-dimensional compact space $\calS \subset \bbR^D$ observed at $V$ spatial locations, along with $J$ covariates, from $N$ individual subjects. For each subject $i = 1, \ldots, N$, let $y_i(\bfs_v) \in \bbR $ denote the image measurement at the spatial location $\bfs_v \in \calS, v = 1, \ldots, V$, and let $\calS_V = \{\bfs_v\}_{v = 1}^{V}$ collect all those locations. Let $\bfx_i = (x_{i1}, \ldots, x_{iJ})^{\top} \in \bbR^J$ denote the $J$-dimensional covariate vector. 
We consider the following spatially varying coefficient model, 
\begin{equation}\label{eq:vc-model}
y_i(\bfs_v) = \sum_{j=1}^{J} x_{ij} \beta_j(\bfs_v) + \alpha_i(\bfs_v) + \epsilon_i(\bfs_v), \quad i = 1, \ldots, N, v = 1, \ldots, V,
\end{equation}
where $\beta_j(\bfs_v) \in \bbR$ characterizes the main effect of the $j$th covariate $x_{ij}$ common for all subjects $i$, $\alpha_i(\bfs_v) \in \bbR$ characterizes the individual deviation from the common main effect that is specific to subject $i$, and $\epsilon_i(\bfs_v) \in \bbR$ is the measurement error that is assumed to have mean zero, variance $\sigma^2(\bfs_v) > 0$, and $\epsilon_i(\bfs_v)$ is independent of $\epsilon_{i'}(\bfs_{v'})$ whenever $i \neq i'$ or $v \neq v'$ for $i, i' \in \{1, \ldots, N\}$ and, $v, v' \in \{1, \ldots, V\}$. 

In model \eqref{eq:vc-model}, we view $\beta_j(\bfs_v), \alpha_i(\bfs_v), \sigma^2(\bfs_v)$ as the realizations of the functions $\beta_j(\bfs), \alpha_i(\bfs)$, $\sigma^2(\bfs)$ at the set of spatial locations $\{\bfs_v\}_{v = 1}^{V}$. In addition, we note that the individual deviation $\alpha_i(\bfs_v)$ plays a similar role as the random effect term in the spatially varying coefficient model in \citet{zhu2014spatially, li2020sparse}. However, a key difference is that in their works, $\alpha_i(\bfs_v)$ are assumed to be copies of a stochastic process, whose covariance structure is to be estimated, whereas in our context, each $\alpha_i(\bfs_v)$ is treated as a realization of a deterministic function, and that function itself is to be estimated directly. Finally, since the error variance $\sigma^2(\bfs_v)$ is also a function of $\bfs_v$, we can flexibly capture the spatial heterogeneity with respect to the degree of variation in the measurement error. 

Write $\bfbeta(\bfs) = (\beta_1(\bfs), \ldots, \beta_J(\bfs))^{\top} \in \bbR^J$, and $\bfalpha(\bfs) = (\alpha_1(\bfs), \ldots, \alpha_N(\bfs))^{\top} \in \bbR^{N}$. We outline some requirements for $\bfbeta(\bfs), \bfalpha(\bfs), \sigma^2(\bfs)$ to ensure the identifiability and interpretability of model \eqref{eq:vc-model}. We give more rigorous definitions in \Cref{sec:theory}. Specifically, we require that the functions $\bfbeta(\bfs), \bfalpha(\bfs), \sigma^2(\bfs)$ are all \emph{piecewise} smooth with a finite number of continuous components. In addition, we impose some form of sparsity, in the sense that, for each covariate $x_j$, there are image regions where the main effect equals zero, and the absolute value of the nonzero main effects have a positive constant lower bound. Similar conditions have been adopted in the imaging analysis literature \citep[e.g.,][]{zhu2014spatially, sun2017store, li2020sparse}.

\subsection{Approximation via deep neural networks}
\label{sec:estimation}

We propose to approximate the functions $\bfbeta(\bfs), \bfalpha(\bfs), \sigma^2(\bfs)$ in model \eqref{eq:vc-model} using deep neural networks. Specifically, we consider a general $L$-layer feed-forward neural network, where the dimension of the input layer is $K_0$, the dimension of the $\ell$th hidden layer is $K_\ell$, $\ell = 1, \ldots, L$, and the dimension of the output layer is $K_{L+1}$. That is, the $\ell$th hidden layer has $K_{\ell}$ hidden nodes, which is often set to $K_1 = \ldots = K_L = K$. For a $K_0$-dimensional vector input $\bfs \in \bbR^{K_0}$, the function produced by this neural network is of the form, 
\begin{equation} \label{eq:nn-model0}
\aleph(\bfs \mid \bftheta) = \bfW_L \bfphi_L \Big( \cdots \bfW_1 \bfphi_1 \big( \bfW_0 \bfs + \bfb_0 \big) + \bfb_1 \cdots \Big) + \bfb_L,
\end{equation}
where $\bfW_{\ell} \in \bbR^{K_{\ell+1} \times K_{\ell}}$ and $\bfb_{\ell} \in \bbR^{K_{\ell+1}}$ are the weight and bias parameters from the $\ell$th hidden layer to the $(\ell+1)$th hidden layer, $\ell = 0, \ldots, L$, and $\bftheta = \{\bfW_{\ell}, \bfb_{\ell}\}_{\ell=0}^{L}$ collects all the parameters. The function $\bfphi_{\ell}(\cdot)$ is a vector-valued nonlinear activation function, where the activation is applied to each element of the $K_{\ell}$-dimensional input vector, and the output is another $K_{\ell}$-dimensional vector, $\ell = 1, \ldots, L$. Common choices of the activation include the sigmoid function $\{1 + \exp(u)^{-1}\}^{-1}$, and the rectified linear unit (ReLU) function $\max(0, u)$. 

\begin{figure}[t!]
\centering
\includegraphics[width=\linewidth]{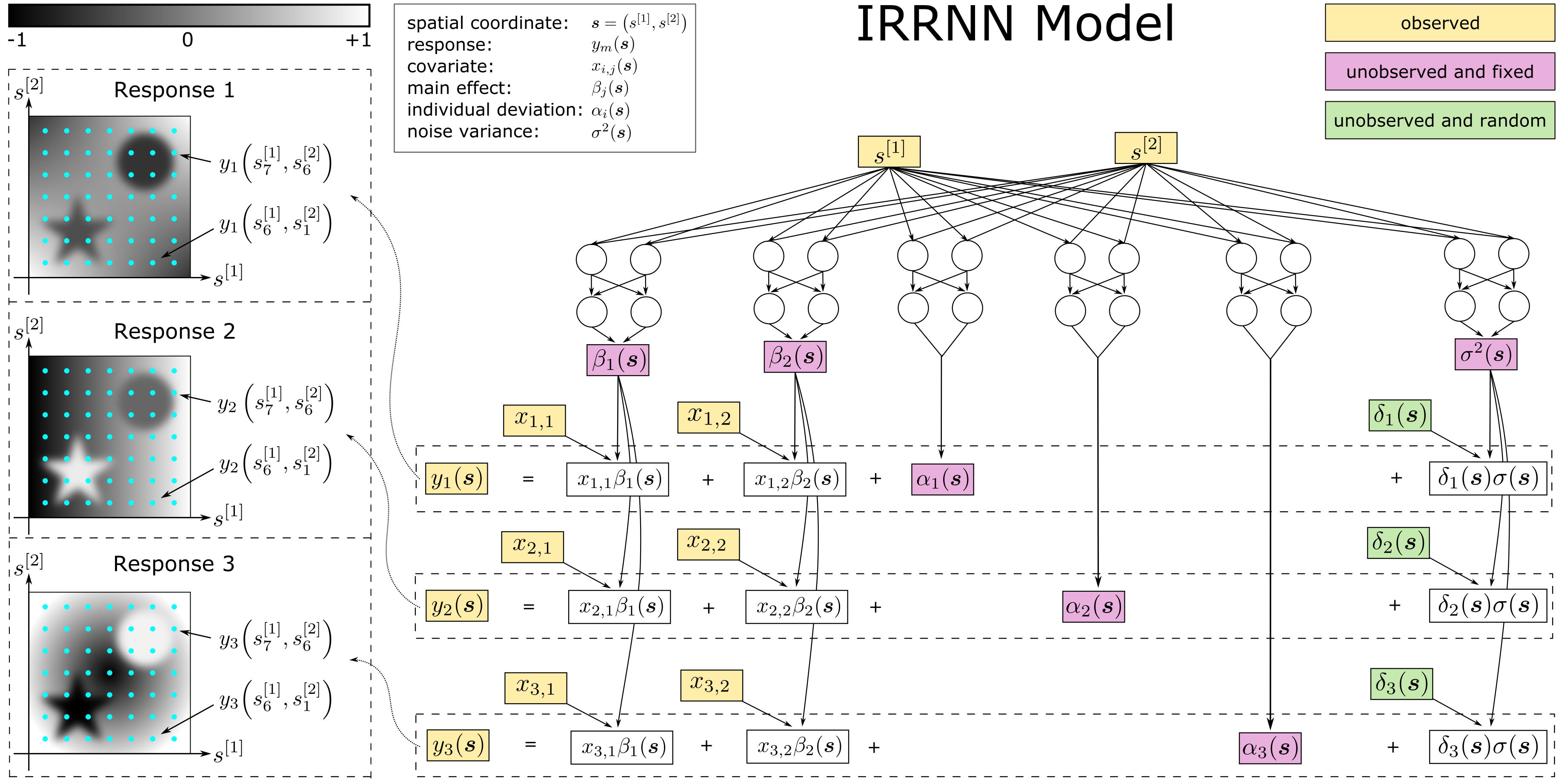} 
\caption{
  Graphical illustration of the proposed method.
  In this example, the data contain $J=2$ covariates and $N=3$ images
  of dimension $D=2$, observed on a grid of $V = 7 \times 7$ voxels.
}
\label{fig:SIR-NN}
\end{figure}

We then approximate $\bfbeta(\bfs), \bfalpha(\bfs), \sigma^2(\bfs)$ using the neural network in \eqref{eq:nn-model0} as, 
\begin{equation} \label{eq:NN_model}
  \bfbeta(\bfs)
  \approx
  \aleph_\beta(\bfs \mid \bftheta_\beta)
  , \quad
  \bfalpha(\bfs)
  \approx
  \aleph_\alpha(\bfs \mid \bftheta_\alpha)
  , \quad
  \log\{\sigma^2(\bfs)\}
  \approx
  \aleph_\sigma(\bfs \mid \bftheta_\sigma).
\end{equation}
The output of the neural networks in \eqref{eq:NN_model}, i.e., $\aleph_\beta(\bfs \mid \bftheta_\beta)$, $\aleph_\alpha(\bfs \mid \bftheta_\alpha)$, and $\aleph_\sigma(\bfs \mid \bftheta_\sigma)$, is of dimension $J$, $N$, and $1$, respectively. That is, $\aleph_\beta(\bfs \mid \bftheta_\beta) = [\aleph_{\beta_1}(\bfs \mid \bftheta_{\beta_1}), \ldots, \aleph_{\beta_J}(\bfs \mid \bftheta_{\beta_J})]^\top$ is a concatenation of $J$ neural networks, and $\aleph_\alpha(\bfs \mid \bftheta_\alpha) = [\aleph_{\alpha_1}(\bfs \mid \bftheta_{\alpha_1}), \ldots$, $\aleph_{\alpha_N}(\bfs \mid \bftheta_{\alpha_N})]^\top$ is a concatenation of $N$ neural networks. The input of the neural networks in \eqref{eq:NN_model} is the $D$-dimensional spatial coordinate, $\bfs_v \in \calS \subset \bbR^D, v = 1, \ldots, V$, which consists of the same fixed design points for all neural networks. In that regard, the model fitting with \eqref{eq:NN_model} is similar to classical regression with fixed-design predictors. For a 3-dimensional image, the input dimension is $D = 3$. \Cref{fig:SIR-NN} gives a graphic illustration of the architecture of our proposed method.

There are several advantages of using \eqref{eq:NN_model} to estimate the unknown functions in model \eqref{eq:vc-model}. First, the neural network is highly flexible in capturing complex patterns of the spatially varying coefficient functions. Second, the interpretation of the functions $\bfbeta(\bfs), \bfalpha(\bfs), \sigma^2(\bfs)$ remains the same as in the classical varying coefficient model. We do not need to interpret the individual parameters $\bftheta_\beta, \bftheta_\alpha, \bftheta_\sigma$ of those fitted neural nets. Moreover, to achieve sparsity to further facilitate the interpretation, we introduce hard thresholding to the main effect function $\bfbeta(\bfs)$, but avoid imposing sparsity on $\bftheta_\beta$. Finally, the effective sample size in \eqref{eq:NN_model} becomes the number of voxels, no longer the number of subjects. For brain images with typical resolutions, the number of voxels is easily in the order of hundreds of thousands or millions, which provides a sufficiently large effective sample size to train the deep neural networks in \eqref{eq:NN_model}. In addition, coupled with low-dimensional input variables, our model fitting is relatively robust to the architecture of the neural network used, such as the number of hidden layers and hidden nodes.   

\subsection{Estimation algorithm}
\label{sec:algorithm}

We devise a three-step algorithm to estimate the functions $\bfbeta(\bfs), \bfalpha(\bfs), \sigma^2(\bfs)$ in model \eqref{eq:vc-model}. We also impose sparsity on the function $\bfbeta(\bfs)$ to further improve the interpretability.  

In Step 1, we estimate the main effect function $\bfbeta(\bfs)$. Write $\bfy(\bfs_v) = (y_1(\bfs_v), \ldots, y_N(\bfs_v))^\top$ $\in \bbR^N$, $v = 1, \ldots, V$, and $\bfX = (\bfx_1, \ldots, \bfx_N)^{\top} \in \bbR^{N \times J}$. We first obtain an initial estimator of the error variance from the mean squared residuals of massive univariate regressions, 
\begin{equation*}
\tilde{\sigma}_v^2 = N^{-1} \big\| \big[ \bfI_N - \bfX (\bfX^\top \bfX)^{-1} \bfX^\top \big] \bfy (\bfs_v) \big\|_2^2, \quad v = 1, \ldots, V,
\end{equation*}
where $\bfI_N$ is the $N \times N$ identity matrix. 

We next estimate the main effect function, by fitting the neural network \eqref{eq:nn-model0}, 
\begin{equation} \label{eq:loss-main-eff}
\hat{\bftheta}_\beta = \underset{\bftheta_\beta}{\arg \min} \sum_{v=1}^V \big\| \bfy(\bfs_v) - \bfX \aleph_\beta(\bfs_v \mid \bftheta_\beta) \big\|^2 \tilde{\sigma}_v^{-2} 
+ \lambda \| \aleph(\bfs_v \mid \bftheta_\beta) \|_1,
\end{equation}
where $\lambda > 0$ is an $L_1$ regularization parameter. We carry out the optimization in \eqref{eq:loss-main-eff} by mini-batch stochastic gradient descent \citep[SGD]{bottou2010large}, where, in our model, a mini-batch of samples in SGD corresponds to a subset of the voxels in $\calS_V$. We denote the resulting estimator from \eqref{eq:loss-main-eff} by $\tilde{\bfbeta}(\bfs) = \aleph_\beta(\bfs | \hat{\bftheta}_\beta)$. 

Next, we apply hard thresholding to the fitted main effect function, i.e., 
\begin{equation} \label{eq:threshold}
\hat{\bfbeta}(\bfs_v) = \tilde{\bfbeta}(\bfs_v) \otimes \bbI \left\{ \big| \tilde{\bfbeta}(\bfs_v) \big| > \bfeta \right\},
\end{equation}
where $\bbI$ is the indicator function, $\otimes$ is the Kronecker product, all operations are  element-wise, and $\bfeta = (\eta_1, \ldots, \eta_J)^{\top} \in \bbR^{J}$ is the covariate-specific thresholding level. The use of the $L_1$ regularization in \eqref{eq:loss-main-eff} is to improve the empirical performance of the estimator, and the $L_0$ regularization in \eqref{eq:threshold} is to induce sparsity and facilitate selection. This procedure is similar to the thresholded Lasso \citep{zhou2010thresholded}.

In Step 2, we estimate the individual deviation function $\bfalpha(\bfs)$, by fitting \eqref{eq:nn-model0}, 
\begin{equation} \label{eq:loss-indiv-eff}
\hat{\bftheta}_\alpha = \underset{\bftheta_\alpha}{\arg \min} \sum_{v=1}^V \big\| \bfy(\bfs_v) - \bfX \hat{\bfbeta}(\bfs_v) - \aleph_\alpha(\bfs_v \mid \bftheta_\alpha) \big\|^2 \tilde{\sigma}_v^{-2},
\end{equation}
where the optimization is done by the mini-batch SGD again. We denote the resulting estimator from \eqref{eq:loss-indiv-eff} by $\hat{\bfalpha}(\bfs_v) = \aleph_\alpha(\bfs_v | \hat{\bftheta}_\alpha)$.

In Step 3, we estimate the error variance function $\sigma^2(\bfs)$. We first update the mean squared residual estimator based on the estimated $\hat{\bfbeta}(\bfs)$ and $\hat{\bfalpha}(\bfs)$ as, 
\begin{equation*}
\bar{\sigma}^2_v = N^{-1} \big\| \bfy(\bfs_v) - \bfX \hat{\bfbeta}(\bfs_v) - \hat{\bfalpha}(\bfs_v) \big\|_2^2, \quad v = 1, \ldots, V.
\end{equation*}
We then estimate the error variance function $\sigma^2(\bfs)$, by fitting the neural network \eqref{eq:nn-model0},  
\begin{equation} \label{eq:loss-noise-var}
\hat{\bftheta}_\sigma = \underset{\bftheta_\sigma}{\arg \min} \sum_{v=1}^V \big\| \bar{\sigma}^2_v - \exp\{\aleph_\sigma(\bfs_v \mid \bftheta_\sigma)\} \big\|_2^2,
\end{equation}
where the optimization is done again by the mini-batch SGD. We denote the resulting estimator from \eqref{eq:loss-noise-var} by $\hat{\sigma}^2(\bfs_v) = \aleph_\sigma(\bfs_v | \hat{\bftheta}_\sigma)$. 

Unlike iterative algorithms, the procedure above only goes through each step once, as, based on our numerical experiments, the estimate does not change much after the first iteration. The one-step estimation procedure also simplifies the subsequent theoretical analysis. Similar one-step estimation methods have been proposed before. For instance, \citet{zhu2014spatially} proposed a one-step estimator that sequentially estimates the main effect, individual deviation, and error variance, followed by smoothing and selection of the main effect estimate.

Our method involves two tuning parameters, the regularization parameter $\lambda$ in \eqref{eq:loss-main-eff}, and the thresholding parameter $\bfeta = \{\eta_j\}_{j=1}^p$ in \eqref{eq:threshold}. To determine $\lambda$, we first adopt voxel-wise Lasso and identify $\lambda_v$ that yields the smallest cross-validation error for that voxel. We then set $\lambda$ as the median of $\{\lambda_v\}_{v=1}^{V}$. To determine $\eta_j$, we first carry out the mass univariate analysis, and choose $\eta_j$, such that $\sum_{v=1}^V \bbI\{
|\tilde{\beta}_j(\bfs_v)| > \eta_j\} = \sum_{v=1}^V \bbI\left\{Z_{j}(\bfs_v) > z_{1-a/2}\right\}$, where $Z_{j}(\bfs_v)$ is the $Z$-statistic at location $\bfs_v$ and $z_{1-\alpha/2}$ is the $(1-a/2)$th quantile of the standard normal distribution, $\alpha = 0.05$, and $j = 1, \ldots, J$. 

Regarding the architecture and implementation of the deep neural network employed in \eqref{eq:nn-model0}, we choose $L = 4$ hidden layers, and for each layer, we choose $K_1 = \ldots = K_L = K = 64$ hidden nodes. Details of the other hyperparameters are described in the Supplementary Materials. Since the dimension of the input in our setting is small, usually only $D=2$ or $3$, while the effective sample size, i.e., the number of voxels, is sufficiently large, we have found that the results are not overly sensitive to the choice of $L$ and $K$. (See \Cref{sec:sensitivity} for details.)

%% file: theory.tex
\section{Theory}
\label{sec:theory}

In this section, we establish the theoretical properties of the proposed method. We first define the function class for the nonparametric functions in our model. We next present the set of regularity conditions, under which we derive the error bounds and establish the estimation and selection consistency.

\subsection{Function class}
\label{sec:funclass}

As is common for any nonparametric theoretical analysis, we first explicitly define the class of candidate functions for our method. We follow the general framework of \citet{bauer2019deep, schmidt2020nonparametric}, and begin with the definition of smooth generalized hierarchical interaction model. We then extend the notion to the concept of the \emph{piecewise} smooth generalized hierarchical interaction model. Finally, we define the candidate neural network function class for our proposed method. 

\begin{definition}[H\"{o}lder smoothness]
\label{def:holder}
For $D \in \bbN_+$, $C \in \bbR_+$, and $P = P' + P'' \in \bbR_+$, where $P' = \lceil P - 1 \rceil$, i.e., the smallest integer no smaller than $P-1$, a function $f: \bbR^D \to \bbR$ is said to be smooth with H\"{o}lder index $P$, or simply $P$-smooth, if for every $(p_1, \ldots, p_D) \in \bbN_0^D$ with $\sum_{d=1}^D p_d = P'$, we have: (a) the partial derivative $\{ \partial^{P'} / (\partial d_1^{p_1} \cdots \partial d_D^{p_D}) \} f$ exists; (b) for any $\bfu \in \bbR^D$, $\{\partial^{P'} / (\partial u_1^{p_1} \cdots \partial u_D^{p_D}) \} f(\bfu) < C$; and (c) for all $\bfu, \bfu' \in \bbR^D$, $\big| {\{\partial^{P'} / (\partial u_1^{p_1} \cdots \partial u_D^{p_D}) \}} f(\bfu) - \{\partial^{P'} / (\partial u_1^{p_1} \cdots \partial u_D^{p_D}) \} f(\bfu') \big| \leq C \| \bfu - \bfu' \|_2^{P''}$. 
\end{definition}

\begin{definition}[Smooth generalized hierarchical interaction model]
\label{def:ghim} 
For $D, D', L', K' \in \bbN_+$ and $P \in \bbR_+$, a function $f: \bbR^D \to \bbR$ is called a $K'$-component $P$-smooth generalized hierarchical interaction model of depth $L'$ and order $D'$, if it satisfies:
\begin{enumerate}[(a)]
\item When $L'=0$, there exist $\bfa_d \in \bbR^{D}$, $d=1, \ldots, D'$, and $g: \bbR^{D'} \to \bbR$, such that $f(\bfu) = g(\bfa_1^\top \bfu, \ldots, \bfa_{D'}^\top \bfu)$, where $g$ is $P$-smooth, and $K'=1$.

\item When $L'>0$, there exist $g_k: \bbR^{D'} \to \bbR$, and $h_{k,d}: \bbR^{D} \to \bbR$, such that $f(\bfu) = \sum_{k=1}^{K'} g_k(h_{k, 1}(\bfu), \ldots, h_{k, D'}(\bfu))$, where $g_k$ is $P$-smooth, and $h_{k,d}$ is a $K'$-component $P$-smooth generalized hierarchical interaction model of depth $L'-1$ and order $D'$, $k = 1, \ldots, K', d=1, \ldots, D'$,
\end{enumerate}
\end{definition}

\Cref{def:holder} gives a general definition of function smoothness, by bounding the derivatives up to a certain degree. \Cref{def:ghim} describes a family of functions that are constructed by recursion and summation of multiple layers of H\"{o}lder-smooth functions defined in \Cref{def:holder}, which covers a wide range of functions  \citep{bauer2019deep}.  

Next, we generalize \Cref{def:ghim}, by allowing the functions to be \emph{piecewise} smooth.

\begin{definition}[Piecewise smooth generalized hierarchical interaction model]
\label{def:ghim-piecewise}
For $D, D'$, $L', K', Q$ $\in \bbN_+$ and $P \in \bbR_+$, a function $f: \bbR^D \to \bbR$ is called a $K'$-component $Q$-sided piecewise $P$-smooth generalized hierarchical interaction model of depth $L'$ and order $D'$, if it satisfies:
\begin{enumerate}[(a)]
\item When $L'=0$, there exist $\bfa_d \in \bbR^D$, $d=1, \ldots, D'$, $g: \bbR^{D'} \to \bbR$, and $\Omega \subset \bbR^D$, such that $f(\bfu) = g(\bfa_1^\top \bfu, \ldots, \bfa_{D'}^\top \bfu) \times \bbI_{\Omega}(\bfu)$, where $g$ is $P$-smooth, $\Omega$ is a $Q$-sided $D$-dimensional polytope, and $K'=1$.

\item When $L'>0$, there exist $g_k: \bbR^{D'} \to \bbR$, $h_{k,d}: \bbR^D \to \bbR$, and $\Omega_k \subset \bbR^D$, such that $f(\bfu) = \sum_{k=1}^{K'} g_k(h_{k, 1}(\bfu), \ldots, h_{k, D'}(\bfu)) \times \bbI_{\Omega_k}(\bfu)$, where $g_k$ is $P$-smooth and $h_{k,d}$ is a $K'$-component $P$-smooth generalized hierarchical interaction model of depth $L'-1$ and order $D'$, and $\Omega_k \subset \bbR^D$ is a $Q$-sided $D$-dimensional polytope, $k = 1, \ldots, K', d=1, \ldots, D'$. 
\end{enumerate}
\end{definition}

\Cref{def:ghim-piecewise} generalizes \Cref{def:ghim}, and describes a family of piecewise smooth functions. This family again covers a wide range of functions. For instance, when $D=2$, the nonzero regions of $\bfbeta$ can be partitioned into a finite number of hexagons, and $\bfbeta$ is Lipschitz continuous inside each partition, $\bfbeta$ is a $6$-sided-piecewise $1$-smooth generalized hierarchical interaction model of depth 1 and order2. When $K=1$ and $\calS \subset \Omega_1$, the function is smooth over the entire spatial domain, and satisfies both \Cref{def:ghim} and \Cref{def:ghim-piecewise}.  

Next, we formally characterize the class of candidate neural network functions. The feed-forward deep neural network model \eqref{eq:nn-model0} we use falls in this class. 

\begin{definition}[Candidate neural network model]
\label{def:nn}
Let $\phi(u) = \{1 + \exp(u)^{-1}\}^{-1}$. For $D$, $D'$, $L'$, $K'$, $Q$, $M \in \bbN_+$ and $\gamma \in \bbR_+$, define $\calG_{D, D', L', K', Q}^{M, \gamma}$ to be the collection of all functions $g: \bbR^D \to \bbR$ that satisfy:
\begin{enumerate}[(a)]
\item When $L' = 0$, there exist $|\xi^{[1]}_{m,b,d}|, |\xi^{[2]}_{m,b} |\xi^{[3]}_m| \in (0, \gamma)$, $m = 0, \ldots, M, b = 0, \ldots, 4D'+Q, d = 0, \ldots, D$, such that
\begin{equation*}
g(\bfu) = \sum_{m=1}^M \xi^{[3]}_m \phi \left( \sum_{b=1}^{4D'+Q} \xi^{[2]}_{m,b} \phi \left( \sum_{d=1}^D \xi^{[1]}_{m,b,d} u_d + \xi^{[1]}_{m,b,0} \right) + \xi^{[2]}_{m,0} \right) + \xi^{[3]}_0,
\end{equation*}
and $K'=1$.

\item When $L' > 0$, there exist $g_k \in \calG_{D, D', 0, K', Q}^{M, \gamma}$ and $h_{k, b} \in \calG_{D, D', L'-1, K', Q}^{M, \gamma}$, $k = 1, \ldots, K', b = 1, \ldots, D'$, such that $g(\bfu) = \sum_{k=1}^{K'} g_k(h_{k,1}(\bfu), \ldots, h_{k,D'}(\bfu))$.
\end{enumerate}
\end{definition}

\Cref{def:nn} describes a family of neural network functions, $\calG_{D, D', L', K', Q}^{M, \gamma}$, obtained from recursively composing and taking the sum of two-layer fully connected feed-forward neural networks. The recursion is repeated $L'$ times. At each recursion, the output is the sum of $K'$ two-layer fully connected feed forward neural networks, where each has $M$ hidden nodes in the second hidden layer, $4D'+Q$ hidden nodes in the first hidden layer, and $D'$ nodes in the input layer. Moreover, we take the logistic function as the activation function $\phi$ in \Cref{def:nn}, following \citet{bauer2019deep}. Similar results can be obtained when using the ReLU activation function \citep{schmidt2020nonparametric}.

We remark that, in our theoretical analysis, we study the family of piecewise smooth functions as described in \Cref{def:ghim-piecewise}, which we assume $\bfbeta(\bfs), \bfalpha(\bfs), \sigma^2(\bfs)$ all belong to. We then employ the candidate neural network functions as described in \Cref{def:nn} to approximate the true functions $\bfbeta(\bfs), \bfalpha(\bfs), \sigma^2(\bfs)$. 

We also remark on the constants introduced in \Cref{def:ghim-piecewise} and \Cref{def:nn}. In particular, $P$ reflects the smoothness level of the functions that we study, and $Q$ is the number of sides of the polytope that characterizes the boundary of the piecewise functions. The constant $D$ is the dimension of the observed image, and $D'$ is the order of the generalized hierarchical interaction model. The constants $L'$ and $K'$ are the number of layers and the number of components in the recursive construction of the true functions and the candidate functions. The constant $\gamma$ is the bound of the absolute values of the weight parameters in the neural network functions, and $M$ corresponds to the number of hidden nodes in the recursive construction of the two-layer feed-forward neural networks. In our theoretical analysis, following \citet{bauer2019deep, schmidt2020nonparametric}, we use the same $L', K'$ for the true and candidate functions. We let $M$ and $\gamma$ control the complexity of the neural network model that we use, and allow $M$ and $\gamma$ to diverge along with the number of voxels $V$ and the number of images $N$.

\subsection{Regularity conditions}
\label{sec:conditions}

We first list a set of regularity conditions needed to establish the convergence rate of the estimation error of our method.

\begin{assumption}
\label{assu:smooth}
  For $i = 1, \ldots, N$ and $j = 1, \ldots, J$, suppose the true functions $\beta_j(\cdot)$, $\alpha_i(\cdot)$, and $\sigma^2(\cdot)$ are $K'$-component $Q$-sided piecewise $P$-smooth generalized hierarchical interaction models of depth $L'$ and order $D'$,
  with H\"older smoothness constant $C$.
  In addition, for the corresponding spatial domain $\calS$, there exists a constant $c_{0} \in \bbR_+$, such that $\calS \subset [-c_{0}, c_{0}]^D$.

\end{assumption}

\begin{assumption}
\label{assu:covariate-bound}
  The random covariates $\bfx_1, \ldots, \bfx_N$ are i.i.d., with $\E[\bfx_1] = \bm{0}$, $\E[\|\bfx_1\|_2^4] < c_{1}$, and $\|\Cov(\bfx_1)\|^2_F < c_{2}$ (where $\| \cdot \|_F$ is the Frobenius norm). In addition, there exist constants $c_{3}, c_{4} \in \bbR_+$, such that $\E(\exp[c_{3} \{\bfx_1^\top \bfbeta(\bfs) + \epsilon(\bfs)\}^2]) < c_{4}$, for all $\bfs \in \calS$.
\end{assumption}

\Cref{assu:smooth} places smoothness requirements on the true functions, which is common in the varying coefficient model literature. For instance, \citet{zhu2014spatially} imposed smoothness by requiring the function to be piecewise Lipschitz continuous. \citet{li2020sparse} imposed smoothness by assuming the function to be in a Sobolev space. Moreover, we assume that the domain of those spatially varying functions is contained in a compact set, which is true in practice for most imaging studies, as the measurement boundary of the imaging machine is usually bounded. \Cref{assu:covariate-bound} places the bounds on the moments of the covariates.

\begin{assumption}
\label{assu:nn-family}
The neural network approximation functions belong to the class specified in \Cref{def:nn}, i.e., $\aleph_{\beta_j}(\cdot \mid \bftheta_{\beta_j}) \in \calG_{D,D',L',K',Q}^{M, \gamma_\beta}$, $\aleph_{\alpha_i}(\cdot \mid \bftheta_{\alpha_n}) \in \calG_{D,D',L',K',Q}^{M, \gamma_\alpha}$, and $\aleph_{\sigma}(\cdot \mid \bftheta_\sigma) \in \calG_{D,D',L',K',Q}^{M, \gamma_\sigma}$, for $i=1, \ldots, N$ and $j = 1, \ldots, J$. Moreover, $\gamma_\beta = (NV)^{c_{5}}$ and $\gamma_\alpha = \gamma_\sigma = V^{c_{6}}$ for some constants $c_5, c_6 \in \bbR_+$. Furthermore, there exists a constant $c_{7} > 0$, such that, for any $\bfs \in \calS$, $\|\aleph(\bfs \mid \hat{\bftheta}_\beta)\|_2^2 < c_{7} \log(NV)$, $\|\aleph(\bfs \mid \hat{\bftheta}_\alpha)\|_2^2 < c_{7} \log(V)$, and $\aleph(\bfs \mid \hat{\bftheta}_\sigma)^2 < c_{7} \log(V)$.
\end{assumption}

\begin{assumption}
\label{assu:nn-estimator}
There exists a constant $c_{8} > 0$ independent of $N$, such that the $L_1$ penalty parameter $\lambda$ in \eqref{eq:loss-main-eff} satisfies $\lambda \leq c_{8} V^{-1}$.  
\end{assumption}

\Cref{assu:nn-family} specifies the candidate neural network functions that we use to approximate the true functions in the varying coefficient model. Moreover, it bounds the absolute values of the weight parameters in the neural network functions $\gamma$, as well as the $L_2$ norm of the neural network estimators. \Cref{assu:nn-estimator} places a rate requirement on the $L_1$ regularization parameter $\lambda$ in \eqref{eq:loss-main-eff}. Similar requirements on the $L_1$ penalty have been imposed in the studies of sign consistency of LASSO-based selection.  For example, in \citet{zhao2006model}, the rate of the penalty weight is set to be strictly between $\calO(1)$ and $\calO(n^{-\frac{1}{2}})$. In comparison, the esitmator in \eqref{eq:loss-main-eff} is penalized with a lighter weight, since we further apply hard thresholding afterwards in \eqref{eq:threshold}. 

We next list two additional regularity conditions needed to establish the convergence rate of the selection error of our method. 

\begin{assumption}
\label{assu:main-eff-sparsity}
There exists a constant $c_{9} \in \bbR+$, such that $\inf_{\beta_j(\bfs) \neq 0} |\beta_j(\bfs)| > c_{9}$, and $\mu\big( \{\bfs \in \calS: \beta_j(\bfs) \neq 0\} \big) / \mu(\calS) \ll 1$, for $j = 1, \ldots, J$, where $\mu$ is a Lebesgue measure of the size of a set in the spatial domain $\calS \subset \bbR^D$.
\end{assumption}

\begin{assumption}
\label{assu:selection-threshold}
There exists a constant $c_{10} \in \bbR_+$, such that $\eta_j \to 0$ as $M \to \infty$, and $\eta_j > c_{10} \log(M)^{-1}$, for $j \in 1, \ldots, J$.
\end{assumption}

\Cref{assu:main-eff-sparsity} requires that the main effects are sparse, and that there is a gap between zero and the weakest signal level. \Cref{assu:selection-threshold} requires the hard thresholding level $\eta_j$ in \eqref{eq:threshold} to converge to zero, but at a rate not decaying too fast. In practice, this condition can be achieved by setting a minimum on the thresholding level.

\subsection{Error bounds and consistency}
\label{sec:maintheory}

In our theoretical analysis, we first establish in \Cref{thm:beta-nothreshold} the error bound of the estimator $\tilde{\bfbeta}(\bfs)$ from \eqref{eq:loss-main-eff}, i.e., the estimator before the hard thresholding step \eqref{eq:threshold}. We next establish in \Cref{thm:beta} the error bound of the estimator $\hat{\bfbeta}(\bfs)$ from \eqref{eq:threshold} after hard thresholding. We then obtain in \Cref{thm:alpha-sigma} the error bound of the estimators $\hat{\bfalpha}(\bfs)$ from \eqref{eq:loss-indiv-eff} and $\hat{\sigma}^2(\bfs)$ from \eqref{eq:loss-noise-var}. Finally, in \Cref{thm:num-nodes-grows,thm:num-images-grows}, we express the error bounds when the neural network complexity is a function of the number of voxels and images.

We first derive the error bound, in terms of the $L_2$ estimation error, of the estimator $\tilde{\bfbeta}(\bfs)$ of the main effect function $\bfbeta$ before applying hard thresholding. Recall that $D$ is the dimension of the observed image, $P$ reflects the degree of smoothness of the true function $\bfbeta$, and $M$ is associated with the total number of nodes and thus reflects the complexity of the neural network functions.

\begin{theorem} \label{thm:beta-nothreshold}
There exists a constant $c_{11} \in \bbR_+$ that, for sufficiently large $N$ and $V$,
\begin{equation*} \label{eq:fit-error}
\E\left\{ \frac{1}{V} \sum_{v=1}^V \left\| \tilde{\bfbeta}(\bfs_v) - \bfbeta(\bfs_v) \right\|_2^2 \right\}
\leq c_{11} \left\{ \log(NV)^3 \left( \frac{M}{NV} + M^{-\frac{2P}{D}} \right) + \frac{1}{N} \right\}.
\end{equation*}
\end{theorem}

\Cref{thm:beta-nothreshold} decomposes the mean squared error of $\tilde{\bfbeta}(\bfs)$ into the sum of three terms. The first term corresponds to the estimation error, i.e., the ``variance'' of the estimator. It decreases as the number of subjects $N$ or the number of voxels $V$ increases, i.e., when more data information becomes available, and it increases as the number of nodes in the neural network increases, since the candidate models become more flexible but also more variable. The second term corresponds to the approximation error, i.e., the ``bias'' for the estimator, that is related to the approximation capability of the neural network. This bias decreases as the neural network involves a larger number of nodes. The third term corresponds to the deviation from the true main effect induced by the error in estimating the individual deviation. Since the individual deviation is also spatially correlated, the error cannot be reduced simply by increasing the number of voxels, nor increasing the complexity of the neural network. 

Next, we derive the error bound, in terms of both the $L_2$ estimation error and the $L_0$ sign error, of the estimator $\hat{\bfbeta}(\bfs)$ of the main effect function $\bfbeta$ after applying hard thresholding. The result is built upon \Cref{thm:beta-nothreshold} and Markov's inequality. 

\begin{theorem} \label{thm:beta}
There exist constants $c_{12}, c_{13} \in \bbR_+$ that, for sufficiently large $N$ and $V$,
\begin{enumerate}[(a)]
\item for the $L_2$ estimation error,
\begin{align*}
\E  \left\{ \frac{1}{V} \sum_{v=1}^V \left\| \hat{\bfbeta}(\bfs_v) - \bfbeta(\bfs_v) \right\|_2^2 \right\} 
\leq c_{12} \left\{ \log(NV)^5 \left( \frac{M}{N V} + M^{-\frac{2P}{D}} \right) + \frac{\log(N)^2}{N} \right\};
\end{align*}

\item for the $L_0$ sign error,
\begin{align*}
\E \left[ \frac{1}{V} \sum_{v=1}^V \left\| \sign\big\{ \hat{\bfbeta}(\bfs_v) \big\} - \sign\big\{ \bfbeta(\bfs_v) \big\} \right\|_0 \right] 
\leq c_{13} \left\{ \log(NV)^5 \left( \frac{M}{N V} + M^{-\frac{2P}{D}} \right) + \frac{\log(N)^2}{N} \right\}.
\end{align*}
\end{enumerate}
\end{theorem}

\Cref{thm:beta} obtains an $L_2$ estimation error bound for $\hat{\bfbeta}(\bfs)$ that is the same as that of $\tilde{\bfbeta}(\bfs)$ in \Cref{thm:beta-nothreshold}, up to a logarithmic factor. It implies that  $\hat{\bfbeta}(\bfs)$ is a consistent estimator of $\bfbeta(\bfs)$. Moreover, the $L_0$ sign error bound is a weighted average between the false positive rate, the false negative rate, and the false sign flipping rate, where the weights depend on the true proportion of the three signs in the main effect function. This $L_0$ sign error bound implies the region selection consistency. That is, let $\mathcal{R}_{j} = \{\bfs_v: \beta_j(\bfs_v) \neq 0, v = 1,\ldots, V\}$ represent the true activation region for covariate $j$, and  $\hat{\mathcal{R}}_{j} = \{\bfs_v: \hat{\beta}(\bfs_v) \neq 0, v = 1,\ldots, V\}$ an estimate of $\mathcal{R}_j$. Then $\hat{\mathcal{R}}_{j}$ converges to $\mathcal{R}_j$ with probability one. To obtain both the estimation and selection consistency, we need the number of images $N$ to go to infinity, with $M$ increasing at a rate slower than $\mathcal{O}(NV)$, while the number of voxels $V$ can be either fixed or diverge to infinity. (See \Cref{thm:num-nodes-grows} and \Cref{thm:num-images-grows} below for a detailed discussion of the relations between $M$, $N$, and $V$ that achieve the best convergence rate.)

Next, we derive the error bound, in terms of the $L_2$ estimation error, of the estimators $\hat{\bfalpha}(\bfs)$ and $\hat{\sigma}^2(\bfs)$ of the individual deviation function $\bfalpha(\bfs)$ and the error variance $\sigma^2(\bfs)$. 

\begin{corollary}
\label{thm:alpha-sigma}
There exist constants $c_{14}, c_{15} \in \bbR_+$ that, for sufficiently large $N$ and $V$,
\begin{enumerate}[(a)]
\item for the $L_2$ estimation error of the individual deviation function estimator, 
\begin{align*}
\E\left\{ \frac{1}{N V} \sum_{v=1}^V \left\| \hat{\bfalpha}(\bfs_v) - \bfalpha(\bfs_v) \right\|_2^2 \right\}
\leq c_{14} \left\{ \log(NV)^5 \left( \frac{M}{N V}+ M^{-\frac{2P}{D}} \right) + \frac{\log(N)^2}{N} + \frac{\log(V)^3 M}{V} \right\}.
\end{align*}

\item for the $L_2$ estimation error of the error variance function estimator,
\begin{align*}
\E\left\{ \frac{1}{V} \sum_{v=1}^V \left\| \hat{\sigma}^2(\bfs_v) - \sigma^2(\bfs_v) \right\|_2^2 \right\}
\leq c_{15} \left\{ \log(NV)^5 \left( \frac{M}{N V}+ M^{-\frac{2P}{D}} \right) + \frac{\log(N)^2}{N} + \frac{\log(V)^3 M}{V} \right\}.
\end{align*}
\end{enumerate}
\end{corollary}

\Cref{thm:alpha-sigma} shows the error bounds of $\hat{\bfalpha}(\bfs)$ and $\hat{\sigma}^2(\bfs)$ converge to zero when both the number of images $N$ and the number of voxels $V$ approach infinity. This is slightly different from the error bound convergence of $\hat{\bfbeta}(\bfs)$ as shown in \Cref{thm:beta-nothreshold}, where only $N$ is required to go to infinity, since when the number of images increases, the number of individual deviation functions to be estimated also increases. As such, a sufficient number of voxels is needed for estimating the individual deviations accurately. The same is true for estimating the error variance function.

In \Cref{thm:beta-nothreshold,thm:beta}, the error bounds of $\hat{\bfbeta}(\bfs)$ are functions of the number of images $N$, the number of voxels $V$, and the neural network complexity measure $M$. Next, we re-express the error bounds, first as functions of $N$ and $V$, then as functions of $V$ only. 

\begin{corollary}
\label{thm:num-nodes-grows}
Suppose $M = c_{16} (NV)^{\frac{D}{D+2P}}$ for some constant $c_{16} \in \bbR_+$. Then there exist constants $c_{17}, c_{18} \in \bbR_+$ that, for sufficiently large $N$ and $V$,
\begin{enumerate}[(a)]
\item for the $L_2$ estimation error,
\begin{align*}
\E \left\{ \frac{1}{V} \sum_{v=1}^V \left\| \hat{\bfbeta}(\bfs_v) - \bfbeta(\bfs_v) \right\|_2^2 \right\}
  \leq c_{17} \left\{ \log(NV)^5 (NV)^{-\frac{2P}{2P+D}} + \log(N)^2 N^{-1} \right\};
\end{align*}

\item for the $L_0$ sign error,
\begin{align*}
\E \left[ \frac{1}{V} \sum_{v=1}^V \left\| \sign\big\{ \hat{\bfbeta}(\bfs_v) \big\} - \sign\big\{ \bfbeta(\bfs_v) \big\} \right\|_0 \right] 
  \leq c_{18} \left\{ \log(NV)^5 (NV)^{-\frac{2P}{2P+D}} + \log(N)^2 N^{-1} \right\}.
\end{align*}
\end{enumerate}
\end{corollary}

\Cref{thm:num-nodes-grows} follows directly from \Cref{thm:beta}, by setting the neural network complexity $M$ proportional to $(NV)^{D/(D+2P)} = (NV)^{1/(1+2P/D)}$. The exponent $1/(1+2P/D)$ controls the rate at which $M$ needs to grow as $N$ and $V$ grow, and this exponent goes to zero as $2P/D$ goes to infinity. Intuitively, when the true function $\bfbeta(\bfs_v)$ is highly smooth, i.e., when $2P/D$ is large, a simple model is sufficient to approximate the true function well, and an overly complex model is more likely to induce overfitting. On the other hand, when the true function is highly non-smooth, i.e., when $2P/D$ is small, a more complex model is needed to ensure an accurate approximation. 

\begin{corollary}
\label{thm:num-images-grows}
Suppose $N = c_{19} V^{\frac{2P}{D}}$ for some constant $c_{19} \in \bbR_+$. Then there exist constants $c_{20}, c_{21} \in \bbR_+$ that, for sufficiently large $V$,
\begin{enumerate}[(a)]
\item for the $L_2$ estimation error,
\begin{align*}
\E \left\{ V^{-1} \sum_{v=1}^V \left\| \hat{\bfbeta}(\bfs_v) - \bfbeta(\bfs_v) \right\|_2^2 \right\}
\leq c_{20} \log(V)^5 V^{-\frac{2P}{D}};
\end{align*}

\item for the $L_0$ sign error,
\begin{align*}
\E \left[ \frac{1}{V} \sum_{v=1}^V \left\| \sign\big\{ \hat{\bfbeta}(\bfs_v) \big\} - \sign\big\{ \bfbeta(\bfs_v) \big\} \right\|_0 \right] 
\leq c_{21} \log(V)^5 V^{-\frac{2P}{D}}.
\end{align*}
\end{enumerate}    
\end{corollary}

\Cref{thm:num-images-grows} further simplifies the rate in \Cref{thm:num-nodes-grows}, by expressing the number of images $N$ as a function of the number of voxels $V$, where $N$ is proportional to $V^{-\frac{2P}{D}}$, to achieve the best error bound in our setting. We note that our error bound is comparable to the minimax error bound $N^{-\frac{2P}{2P+D}}$ in the classical nonparametric regression setting \citep{stone1982optimal}.

%% file: simulation.tex
\section{Simulations} 
\label{sec:simulations}

In this section, we first lay out the setup of our simulation experiments. We then present the empirical performance of our proposed method, and compare it to some alternative solutions. We also carry out a sensitivity analysis to evaluate the robustness of our method with respect to the architecture of the deep neural network.

\FloatBarrier
\begin{figure}[!ht]
\centering
  \caption{
    Image slices of the spatially varying functions
    of the true main effect, noise variance,
    an individual deviation,
    and a response image,
    along with the estimated main effects by various methods.
  }
\begin{subfigure}[b]{0.9\textwidth}
\centering
\includegraphics[width=4in]{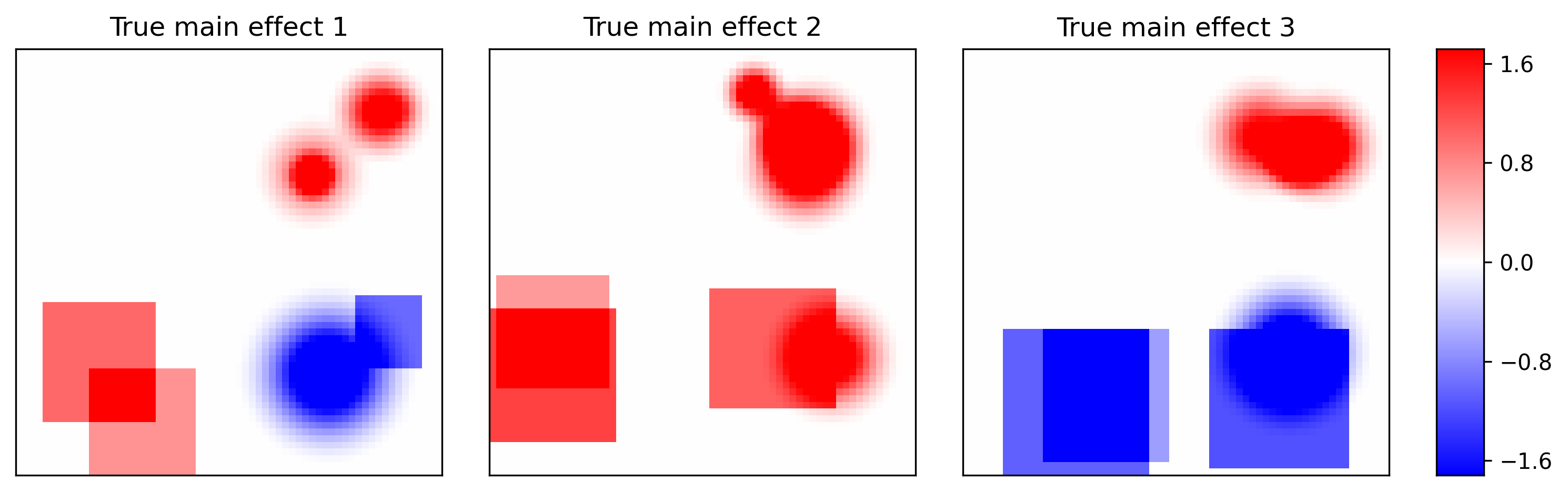} \\
\hspace*{0.800in} \includegraphics[width=3.582in, left]{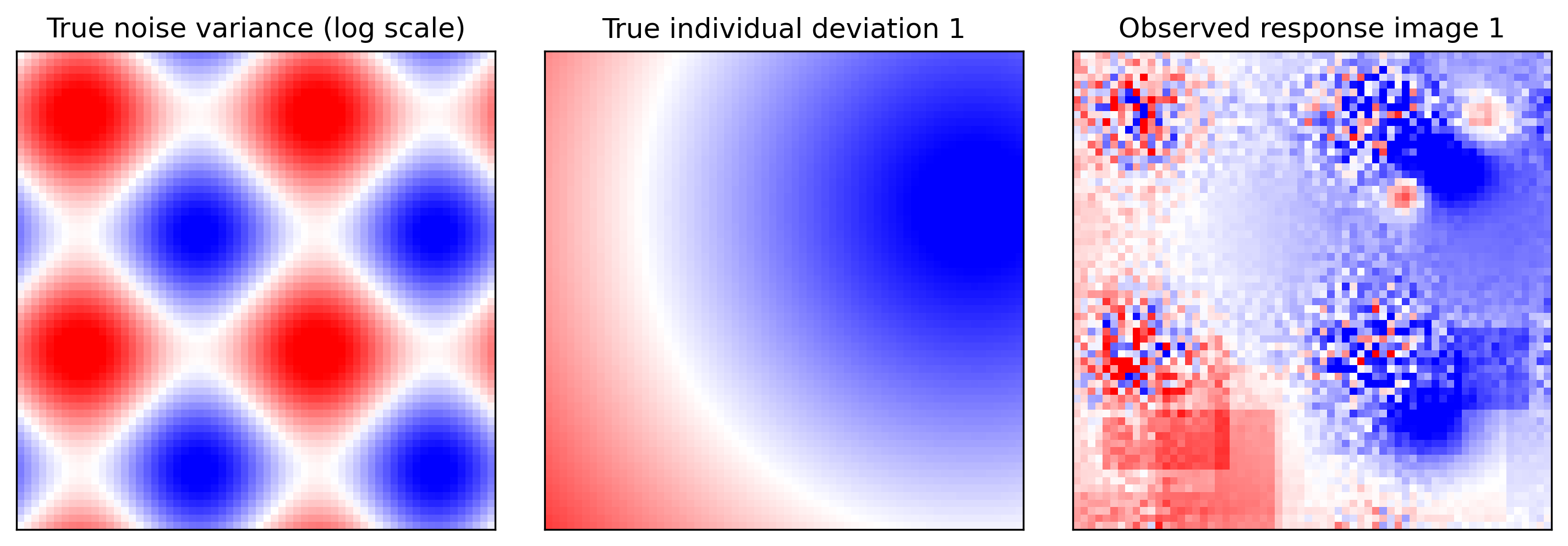} \\
\caption{\label{fig:simu-truth} True spatially varying coefficient functions and a response image.}
\end{subfigure}
\begin{subfigure}[b]{0.9\textwidth}
\centering
\includegraphics[width=4in]{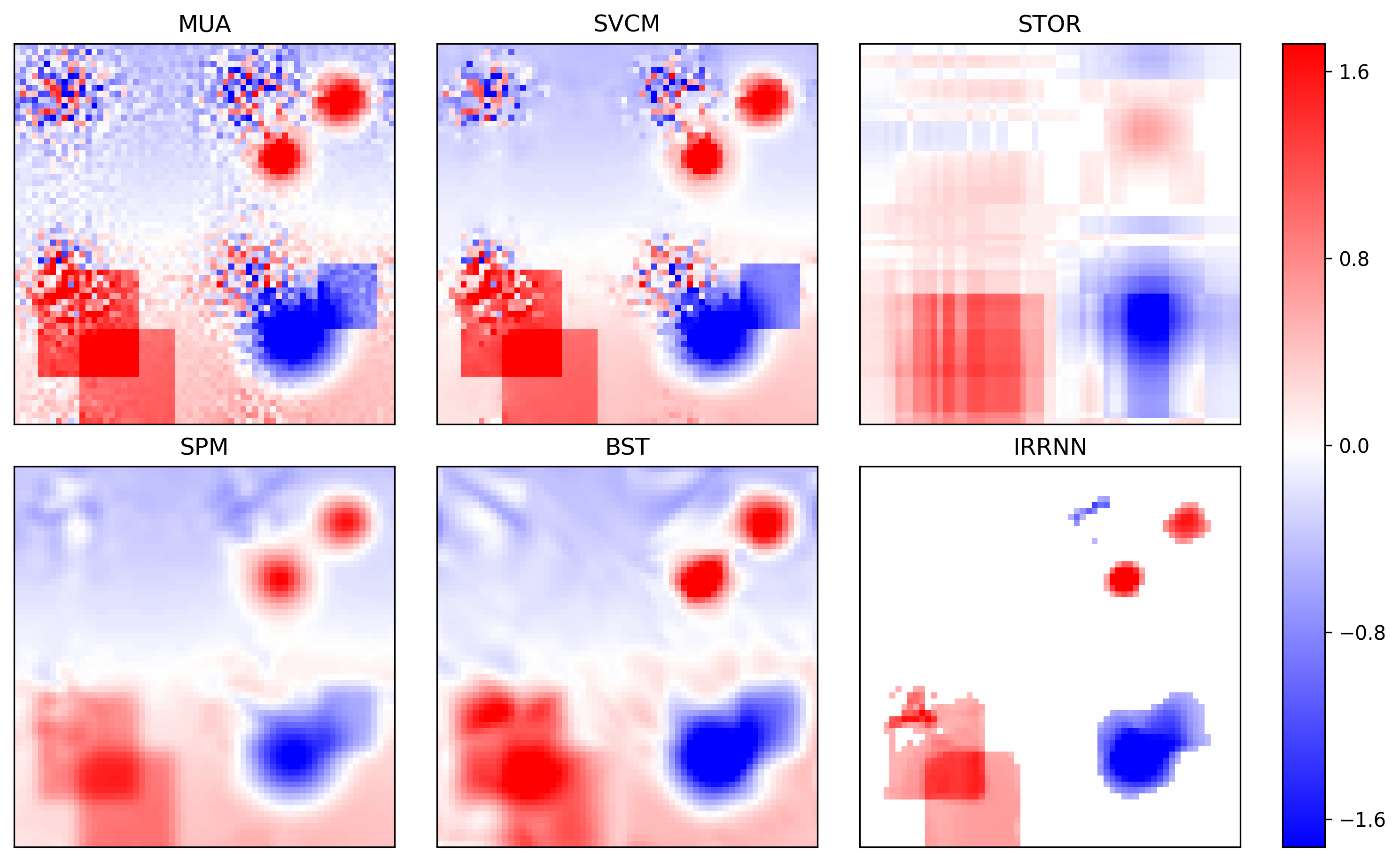}
\caption{\label{fig:simu-estimates} Estimated spatially varying coefficient functions by various methods.}
\end{subfigure}
\end{figure}

\subsection{Experimental setup}
\label{sec:sim-setup}

We generate the data following model \eqref{eq:vc-model}, with $J=3$ covariates $\bfx_i  = (x_{i1}, \ldots, x_{iJ})^{\top}$, $i = 1, \ldots, N$, independently drawn from a standard normal distribution. We consider two sample sizes, $N = 20$ and $50$. We consider $D=3$-dimensional images, with four image sizes, $16 \times 16 \times 8$, $32 \times 32 \times 8$, $64 \times 64 \times 8$, and $128 \times 128 \times 8$, resulting in $V$ = 2,048, 8,192, 32,768, and 131,072 voxels, respectively. We generate the main effect function $\beta_j(\bfs), j=1, \ldots, J$, the individual deviation function $\alpha_i(\bfs), i = 1, \ldots, N$, and the error variance function $\sigma^2(\bfs)$ from a bounded 3-dimensional rectangular spatial domain $\calS$. We consider different spatial patterns, as illustrated in \Cref{fig:simu-truth}. 

More specifically, for $\beta_j(\bfs)$, we divide the rectangular domain into four blocks, along the first two axes, and consider different spatial patterns in each of the four blocks. For the top-left block, we set the value of the spatially varying function to zero, so this region contains no signal. For the top-right block, we generate two spherical regions that contain signals with strength smoothly diminishing to zero at the boundary. The size and location of the two spherical regions are randomly and independently selected, and the two regions may be mutually exclusive or overlapping. If there is an overlap, the values inside the overlapping region are simply the sum of the values of the two individual regions. For the bottom-left block, we generate two rectangular regions with a constant signal level, resulting in sharp rather than smooth edges. For the bottom-right block, we generate a spatial pattern by mixing those in the top-right block and the bottom-left block. \Cref{fig:simu-truth} top row shows one realization of the generated main effect function. 

For $\alpha_i(\bfs)$, we first randomly select a spatial location as the center and assign a random signal value. We then proportionally increase or decrease the signal value according to the distance to the center. \Cref{fig:simu-truth} bottom left panel shows one realization of the generated individual deviation function. 

For $\sigma^2(\bfs)$, we generate the noise variance in a periodic manner across the three axes by using a random linearly transformed sine function. \Cref{fig:simu-truth} bottom center panel shows one realization of the generated error variance function. Given the variance, we generate the measurement error from two distributions, a standard normal and a $(\chi^2_3 - 3)/\sqrt{6}$ distribution. 

In our simulations, we consider the setting where the signal strength is relatively low, by choosing the ratio of the variances of the main effect, the individual deviation, and the error to $0.2 : 0.5 : 1.0$. This mimics the highly noisy imaging data setting that is often the rule rather than the exception in biomedical imaging studies.  \Cref{fig:simu-truth} bottom right panel shows a two-dimensional slice of a realization of the generated response image.

\FloatBarrier

\subsection{Empirical estimation and selection accuracy}
\label{sec:sim-results}

We compare our method with a number of alternative solutions. These include the massive univariate analysis (MUA), and the massive univariate analysis after pre-smoothing (SPM), in combination with random field theory for selection, which is implemented in the neuroimaging software \texttt{SPM} \citep{friston2003statistical}. Both of these methods are frequently used in the neuroscience community. We also compare our method with the sparse tensor response regression (STOR) of \citet{sun2017store}, the spatially varying coefficient model with jump discontinuities (SVCM) of \citet{zhu2014spatially}, and the bivariate splines over triangulation-based spatially varying coefficient model (BST) of \citet{li2020sparse}. (Since BST is designed for two-dimensional images, we apply it to slices of the three-dimensional image and then stack the two-dimensional estimates). We name our method as image response regression via neural networks (IRRNN).
We evaluate and compare the methods in terms of estimation and selection accuracy. For the estimation accuracy, we report the mean squared error (MSE) between the estimated and the true function. For the selection accuracy, we report the area under the operating characteristic curve (AUC), the false positive error rate (FPR), and the true positive rate (TPR). We repeat each simulation 50 times, and report the median and interquartile range of various criteria across 50 data replications. We have chosen the median because some of the alternative methods can produce unstable estimates. \Cref{tbl:simu} reports the empirical MSE, AUC, FPR and TPR.  

\begin{table}[t!]
\caption{\label{tbl:simu} The estimation and selection accuracy of various methods. The number of images is $N = 20, 50$. The image dimension is $R \times R \times 8$, with $R = 16 ,32, 64, 128$. The reported numbers are in 0.01 units. }
\footnotesize
\setlength{\tabcolsep}{0.5pt}
\renewcommand{\arraystretch}{0.5}
\begin{subtable}[h]{\textwidth}
    \centering
    \setlength\tabcolsep{1.2pt}
    \begin{tabular}{rr*{12}{rl}}
    \toprule
    \multicolumn{2}{c}{}
    & \multicolumn{12}{c}{$\calN(0,1)$ noise}
    & \multicolumn{12}{c}{$(\chi^2_3-3)/\sqrt{6}$ noise} \\
    \cmidrule(lr){3-14}  \cmidrule(lr){15-26}
    $M$ & $R$
    & \multicolumn{2}{c}{MUA} 
    & \multicolumn{2}{c}{SPM} 
    & \multicolumn{2}{c}{BST} 
    & \multicolumn{2}{c}{STOR} 
    & \multicolumn{2}{c}{SVCM} 
    & \multicolumn{2}{c}{IRRNN} 
    & \multicolumn{2}{c}{MUA} 
    & \multicolumn{2}{c}{SPM} 
    & \multicolumn{2}{c}{BST} 
    & \multicolumn{2}{c}{STOR} 
    & \multicolumn{2}{c}{SVCM} 
    & \multicolumn{2}{c}{IRRNN} \\
      \midrule
      \multicolumn{26}{c}{Mean squared error (MSE)} \\
      \midrule
         20 &  16 & 191 & (57) & 61 & (32) & 63 & (31) & 63 & (27) & 157 & (51) & 38 & (22)
                  & 187 & (78) & 66 & (41) & 69 & (43) & 59 & (36) & 153 & (73) & 35 & (19)
      \\ 20 &  32 & 190 & (59) & 59 & (36) & 59 & (34) & 51 & (33) & 161 & (53) & 28 & (15)
                  & 185 & (64) & 63 & (41) & 64 & (41) & 52 & (31) & 155 & (64) & 25 & (18)
      \\ 20 &  64 & 192 & (58) & 58 & (37) & 58 & (35) & 47 & (32) & 160 & (50) & 23 & (19)
                  & 185 & (66) & 62 & (41) & 62 & (42) & 53 & (31) & 158 & (64) & 20 & (18)
      \\ 20 & 128 & 193 & (54) & 58 & (37) & 56 & (36) & 44 & (31) & 161 & (50) & 22 & (16)
                  & 186 & (64) & 62 & (41) & 61 & (42) & 50 & (30) & 157 & (63) & 19 & (19)
      \\ \midrule
         50 &  16 &  66 & (12) & 30 & (10) & 27 & (11) & 28 & (10) & 52  & (12) & 25 & (08)
                  &  65 & (16) & 32 & (14) & 26 & (14) & 30 & (10) & 51  & (15) & 24 & (11)
      \\ 50 &  32 &  65 & (11) & 26 & (11) & 24 & (11) & 24 & (08) & 52  & (10) & 16 & (08)
                  &  66 & (17) & 27 & (13) & 24 & (13) & 26 & (09) & 52  & (14) & 15 & (10)
      \\ 50 &  64 &  65 & (10) & 24 & (11) & 22 & (11) & 23 & (10) & 52  & (11) & 12 & (07)
                  &  66 & (16) & 25 & (13) & 22 & (13) & 22 & (11) & 53  & (14) & 13 & (08)
      \\ 50 & 128 &  64 & (10) & 24 & (10) & 21 & (10) & 22 & (08) & 51  & (11) & 11 & (06)
                  &  67 & (15) & 24 & (14) & 21 & (14) & 21 & (09) & 53  & (13) & 11 & (08)
      \\ \midrule
      \multicolumn{26}{c}{
          Area under the operating characteristic curve (AUC)
      } \\
      \midrule
         20 & 16  & 66 & (08) & 63 & (13) & 64 & (11) & 61 & (12) & 67 & (09) & 65 & (10)
                  & 68 & (07) & 64 & (12) & 66 & (08) & 62 & (14) & 67 & (08) & 64 & (10)
      \\ 20 & 32  & 66 & (08) & 66 & (11) & 66 & (11) & 64 & (15) & 67 & (10) & 69 & (08)
                  & 67 & (07) & 67 & (10) & 68 & (07) & 64 & (12) & 67 & (08) & 70 & (10)
      \\ 20 & 64  & 66 & (07) & 67 & (11) & 67 & (10) & 65 & (12) & 67 & (09) & 70 & (06)
                  & 67 & (07) & 68 & (10) & 69 & (07) & 66 & (09) & 68 & (08) & 71 & (09)
      \\ 20 & 128 & 66 & (08) & 67 & (11) & 68 & (11) & 64 & (09) & 67 & (09) & 70 & (06)
                  & 67 & (07) & 68 & (10) & 69 & (07) & 64 & (11) & 68 & (08) & 71 & (09)
      \\ \midrule
         50 &  16 & 76 & (08) & 74 & (11) & 77 & (09) & 74 & (14) & 77 & (08) & 78 & (09)
                  & 75 & (07) & 74 & (12) & 76 & (08) & 73 & (09) & 77 & (08) & 78 & (09)
      \\ 50 &  32 & 76 & (08) & 77 & (10) & 78 & (09) & 77 & (14) & 77 & (08) & 81 & (06)
                  & 75 & (07) & 77 & (10) & 77 & (09) & 73 & (10) & 76 & (08) & 79 & (08)
      \\ 50 &  64 & 76 & (08) & 78 & (09) & 79 & (09) & 76 & (16) & 78 & (08) & 81 & (05)
                  & 75 & (06) & 78 & (09) & 78 & (08) & 75 & (09) & 77 & (07) & 80 & (08)
      \\ 50 & 128 & 76 & (08) & 79 & (08) & 79 & (09) & 76 & (15) & 78 & (08) & 81 & (06)
                  & 75 & (07) & 78 & (09) & 78 & (09) & 74 & (12) & 77 & (07) & 81 & (08)
      \\ \midrule
      \multicolumn{26}{c}{False positive rate (FPR) } \\
      \midrule
         20 &  16 & 04 & (05) & 00 & (00) & 23 & (20) & 45 & (53) & 08 & (06) & 05 & (04)
                  & 04 & (04) & 00 & (00) & 23 & (14) & 51 & (59) & 07 & (04) & 04 & (04)
      \\ 20 &  32 & 04 & (05) & 00 & (00) & 23 & (21) & 80 & (85) & 08 & (06) & 04 & (04)
                  & 03 & (04) & 00 & (00) & 23 & (14) & 75 & (79) & 07 & (04) & 03 & (04)
      \\ 20 &  64 & 04 & (05) & 00 & (00) & 23 & (21) & 82 & (76) & 08 & (07) & 03 & (04)
                  & 03 & (04) & 00 & (00) & 24 & (15) & 85 & (61) & 08 & (04) & 03 & (04)
      \\ 20 & 128 & 04 & (05) & 00 & (00) & 23 & (22) & 89 & (81) & 08 & (07) & 03 & (04)
                  & 03 & (04) & 00 & (00) & 24 & (15) & 84 & (75) & 08 & (05) & 03 & (04)
      \\ \midrule
         50 &  16 & 04 & (04) & 00 & (00) & 38 & (14) & 41 & (47) & 07 & (06) & 05 & (04)
                  & 05 & (05) & 00 & (00) & 41 & (20) & 56 & (52) & 08 & (06) & 05 & (03)
      \\ 50 &  32 & 03 & (05) & 00 & (00) & 39 & (15) & 86 & (79) & 08 & (06) & 03 & (04)
                  & 04 & (05) & 00 & (00) & 42 & (19) & 63 & (72) & 08 & (06) & 04 & (04)
      \\ 50 &  64 & 03 & (05) & 00 & (00) & 39 & (15) & 85 & (82) & 07 & (06) & 03 & (04)
                  & 04 & (05) & 00 & (00) & 42 & (19) & 90 & (63) & 08 & (06) & 04 & (03)
      \\ 50 & 128 & 04 & (05) & 00 & (00) & 39 & (14) & 69 & (82) & 07 & (06) & 03 & (04)
                  & 04 & (05) & 00 & (00) & 43 & (19) & 79 & (74) & 08 & (06) & 03 & (04)
      \\ \midrule
      \multicolumn{26}{c}{True positive rate (TPR) } \\
      \midrule
         20 &  16 & 21 & (08) & 00 & (01) & 45 & (20) & 69 & (54) & 28 & (11) & 19 & (09)
                  & 23 & (09) & 00 & (01) & 45 & (18) & 71 & (59) & 29 & (12) & 21 & (10)
      \\ 20 &  32 & 22 & (07) & 01 & (02) & 47 & (16) & 85 & (75) & 27 & (10) & 24 & (09)
                  & 24 & (09) & 01 & (02) & 49 & (16) & 81 & (65) & 29 & (14) & 25 & (10)
      \\ 20 &  64 & 22 & (07) & 01 & (01) & 48 & (15) & 91 & (36) & 28 & (10) & 25 & (07)
                  & 23 & (08) & 01 & (02) & 51 & (14) & 95 & (33) & 30 & (13) & 26 & (10)
      \\ 20 & 128 & 22 & (07) & 01 & (01) & 49 & (16) & 93 & (55) & 28 & (10) & 25 & (07)
                  & 24 & (09) & 01 & (01) & 52 & (16) & 92 & (47) & 30 & (13) & 26 & (10)
      \\ \midrule
         50 &  16 & 42 & (09) & 02 & (03) & 73 & (09) & 79 & (43) & 49 & (11) & 36 & (11)
                  & 43 & (07) & 03 & (03) & 73 & (10) & 82 & (28) & 50 & (07) & 39 & (10)
      \\ 50 &  32 & 42 & (10) & 05 & (04) & 75 & (10) & 97 & (39) & 41 & (12) & 42 & (09)
                  & 43 & (08) & 05 & (04) & 75 & (10) & 89 & (28) & 50 & (09) & 44 & (09)
      \\ 50 &  64 & 42 & (10) & 05 & (03) & 76 & (10) & 97 & (44) & 50 & (11) & 43 & (09)
                  & 43 & (08) & 06 & (04) & 76 & (10) & 97 & (23) & 51 & (09) & 45 & (09)
      \\ 50 & 128 & 42 & (10) & 04 & (03) & 78 & (09) & 94 & (46) & 50 & (11) & 45 & (09)
                  & 43 & (08) & 05 & (03) & 77 & (10) & 95 & (36) & 51 & (09) & 46 & (09)
      \\ \bottomrule
\end{tabular}      
\end{subtable}
\end{table}

For the estimation accuracy, our proposed method achieves uniformly the smallest MSE. We make a few additional observations. The advantage of IRRNN is more prominent when the imaging resolution is high. There is a clear trend of decreasing MSE for IRRNN when the number of voxels $V$ increases. This agrees with our expectation, as we turn $V$ as our effective sample size. Moreover, the improvement of IRRNN is more prominent when the number of images $N$ is small than when $N$ is large. This demonstrates the advantage of our method for medical imaging studies with a limited sample size. Finally, our method is robust against a skewed error distribution. In contrast, the alternative methods of SPM, BST and STOR, when the number of images is small, show a clear increase of MSE as the error becomes a chi-squared distribution. 

For selection accuracy, our proposed method achieves the highest AUC, except for when $N$ and $V$ are both the smallest, in which case our AUC is still comparable to the alternative methods. Meanwhile, our method achieves the best overall performance in terms of FPR and TPR. SVCM, BST, and STOR tend to be overly liberal in selection, while SPM tend to be excessively conservative. MUA has the best selection balance among the alternative methods, but in comparison, as $V$ increases, IRRNN achieves higher TPR, without losing control over the FNR. In addition, the performance of IRRNN is robust for both the Gaussian and the chi-squared error distributions. 

\Cref{fig:simu-estimates} shows the estimated main effect function for one data replication by various estimation methods, for $N = 20$ images of dimension $64 \times 64 \times 8$ with a Gaussian error. MUA produces the most noisy estimate, which is expected as it ignores the spatial information. Its estimation accuracy on an individual voxel highly depends on its local noise level, as shown by the contrast of the estimates
in the noisy regions versus the other regions. Next, SVCM, which is based on smoothing the noise in the MUA estimates, eliminates the majority of the estimation errors in the low-noise regions, but most of the fluctuations in the high-noise regions have remained. In contrast, SPM and BST are less susceptible to white noise, but they tend to generate errors of wrinkled patterns, ignore the bias from the gradual transition in the background, and over-smooth the true signals, as shown in the blurry boundaries of the rectangular activation regions in the estimates. Moreover, STOR favors activation regions with rectangular boundaries, due to its treatment of images as low-rank tensors, but it tends to overlook the activation regions with curvy boundaries. Finally, IRRNN successfully detects all the activation regions of various geometric shapes, with the ability to adapt for both smooth and sharp boundaries. It also eliminates most of the noise and generates the cleanest estimate. In summary, IRRNN achieves the most competitive performance compared to the alternative methods, all under a rather limited sample size with a relatively large amount of noise in the data. 

\subsection{Sensitivity analysis}
\label{sec:sensitivity}

To assess the robustness of our method with respect to the architecture of the neural network, we repeat the previous experiments by using network architectures with various depths and widths, from 2 layers $\times$ 16 nodes per layer to 7 layers $\times$ 512 nodes per layer.
After replicating the experiment on each architecture for 50 times, the median and IQR of the estimation MSE are reported in the Supplementary Materials. Estimation accuracy of IRRNN varies little across different DNN architectures when each layer contains at least 64 nodes. Although the MSE becomes higher when the number of nodes per layer is reduced to 32 or less, our method is still substantially more accurate and more stable than the alternative methods. This result shows that our model is relatively robust against the complexity of the neural networks.

\FloatBarrier

%% file: realdata.tex
\section{Brain Imaging Applications} 
\label{sec:realdata}
\FloatBarrier

In this section, we illustrate our method with two neuroimaging applications, the Autism Brain Imaging Data Exchange (ABIDE) study, and the Adolescent Brain Cognitive Development (ABCD) study. Both studies have collected fMRI images from multiple imaging sites, along with clinical and demographic information. We investigate the association between cognitive ability and imaging response, which corresponds to the degree of inter-brain connectedness in ABIDE and brain activity when engaging working memory in ABCD.

\subsection{ABIDE study}

The overarching goal of the ABIDE study is to improve the neurological understanding of autism and its associated cognitive behaviors \citep{di2014autism}. In our data analysis, we use the Phase I data that consists of fMRI images from 19 imaging sites. The imaging data are preprocessed with the usual pipeline of \citet{cpac2013,he2019selective}. The response image is the weighted degrees of network connectivity, which measure each voxel's number of direct connections to other voxels. The resulting imaging dimension is $61 \times 73 \times 61$. The $J=4$ covariates include cognitive ability, autism status, age, and sex. After removing the missing values, the number of subjects is $N = 821$. We apply IRRNN and compare it with MUA, SPM, and SVCM. We are unable to evaluate STOR and BST, as their official software programs report error messages when analyzing this particular dataset. 

We first evaluate the estimation accuracy, by reporting the cross-validation of the mean squared error of the recovered response image.
That is, we apply the methods to the data from one imaging site, then evaluate them on the data from all other imaging sites. \Cref{tab:sitewise-fit} reports the cross-site testing MSE. We sort the experiments by the performance of MUA, which reflects the baseline difficulty of each estimation task. In the majority of the analyses, our method has the best estimation accuracy. In particular, when the estimation task is relatively easy, as indicated by the smaller MSE by MUA, all methods have similar performance. However, as difficulty increases, the differences between the methods are widened, and IRRNN become consistently more accurate than the alternative methods.

\begin{table}[!t]
  \centering
  \caption{
      \label{tab:sitewise-fit}
      Cross-site testing MSE for recovering imaging response
      in the brain fMRI studies.
      Each row represents a single-site analysis,
      where the data from one study site is used for training
      and those from the other sites are used for testing.
  }
  \begin{subtable}[t]{0.45\linewidth}
    \centering
    \subcaption{ABIDE}
    \begin{tabular}[t]{rrrr}
      \toprule
       MUA &  SPM &  SVCM &  IRRNN \\
      \midrule
      1.54 & 1.46 &  1.50 &   \bf{1.43} \\
      1.62 & 1.80 &  \bf{1.59} &   1.65 \\
      1.69 & \bf{1.47} &  1.68 &   1.63 \\
      1.79 & 1.98 &  1.77 &   \bf{1.76} \\
      1.80 & \bf{1.53} &  1.75 &   1.65 \\
      1.83 & \bf{1.53} &  1.79 &   1.68 \\
      1.83 & \bf{1.59} &  1.74 &   1.68 \\
      \bf{1.91} & 2.18 &  1.91 &   1.98 \\
      1.96 & 2.07 &  1.92 &   \bf{1.84} \\
      \bf{1.98} & 2.27 &  1.98 &   2.00 \\
      2.02 & 2.06 &  1.92 &   \bf{1.76} \\
      2.47 & 2.29 &  2.25 &   \bf{1.76} \\
      \bf{2.62} & 2.89 &  2.62 &   2.69 \\
      3.26 & 2.39 &  2.80 &   \bf{2.19} \\
      3.61 & 2.79 &  3.21 &   \bf{1.74} \\
      5.18 & 3.76 &  4.99 &   \bf{3.65} \\
      6.00 & 3.58 &     - &   \bf{3.03} \\
      6.61 & 4.41 &  5.93 &   \bf{3.28} \\
      7.33 & 5.25 &  7.05 &   \bf{5.41} \\
      \bottomrule
    \end{tabular}
  \end{subtable}
  \hfill
  \begin{subtable}[t]{0.45\linewidth}
    \centering
    \caption{ABCD}
    \begin{tabular}[t]{rrrr}
      \toprule
       MUA &   SPM &   SVCM &  IRRNN \\
      \midrule
      2.74 &  2.71 &   2.72 &   \bf{2.69} \\
      2.76 &  2.72 &   2.72 &   \bf{2.68} \\
      2.83 &  2.81 &   2.80 &   \bf{2.78} \\
      2.84 &  2.79 &   2.80 &   \bf{2.74} \\
      2.85 &  2.79 &   2.81 &   \bf{2.75} \\
      2.85 &  2.80 &   2.81 &   \bf{2.74} \\
      2.86 &  2.79 &   2.82 &   \bf{2.73} \\
      2.93 &  \bf{2.88} &   2.91 &   2.89 \\
      2.97 &  2.89 &   2.91 &   \bf{2.76} \\
      3.07 &  \bf{2.92} &   3.02 &   2.97 \\
      3.09 &  2.95 &   2.98 &   \bf{2.83} \\
      3.44 &  3.38 &   3.39 &   \bf{3.31} \\
      3.47 &  3.50 &   \bf{3.47} &   3.50 \\
      3.50 &  3.50 &   3.48 &   \bf{3.42} \\
      3.53 &  3.52 &   \bf{3.51} &   3.54 \\
      3.71 &  3.61 &   3.63 &   \bf{3.58} \\
      4.30 &  3.66 &   3.85 &   \bf{3.02} \\
      5.82 &  4.87 &   5.24 &   \bf{4.01} \\
      7.28 &  5.58 &   6.34 &   \bf{3.63} \\
      113.28 & 73.12 & 113.28 &   \bf{4.16} \\
    \bottomrule
  \end{tabular}
  \end{subtable}
\end{table}

\begin{figure}[t!]
\centering
\begin{subfigure}{0.49\textwidth}
\centering
        \includegraphics[width=\textwidth]{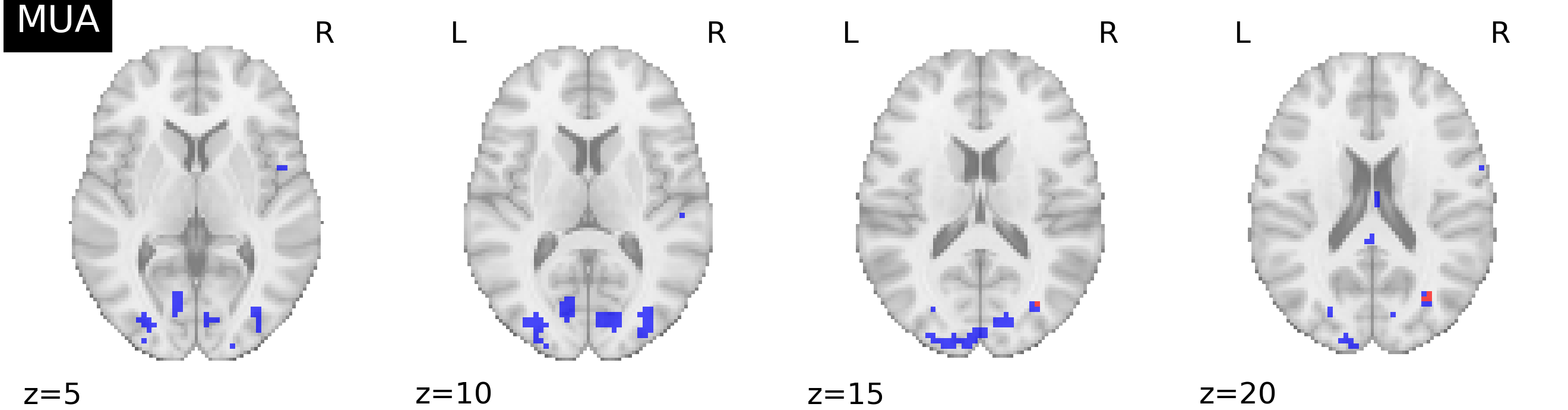}
        \includegraphics[width=\textwidth]{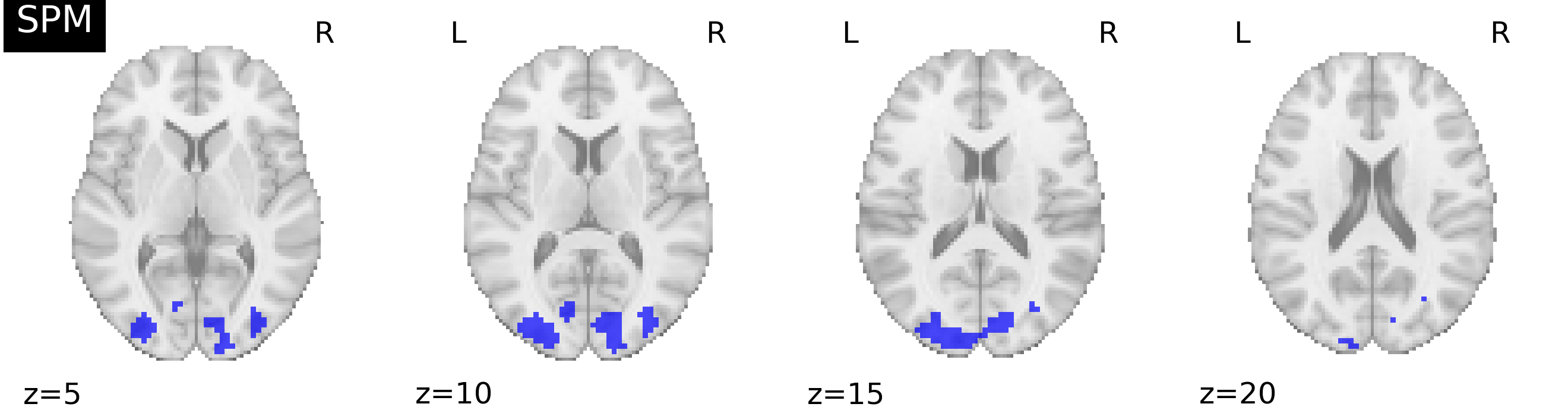}
        \includegraphics[width=\textwidth]{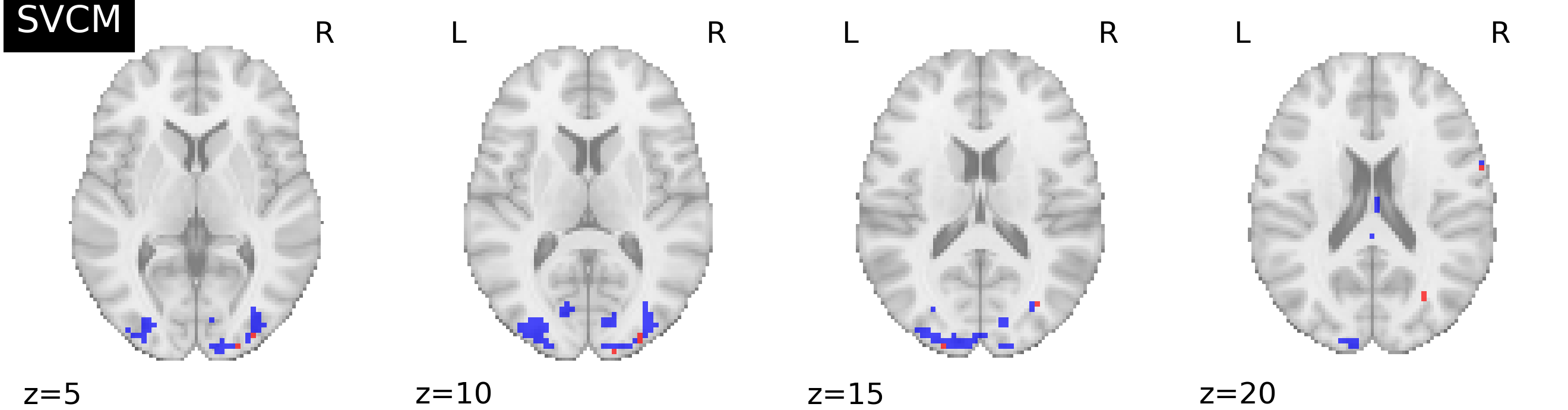}
        \includegraphics[width=\textwidth]{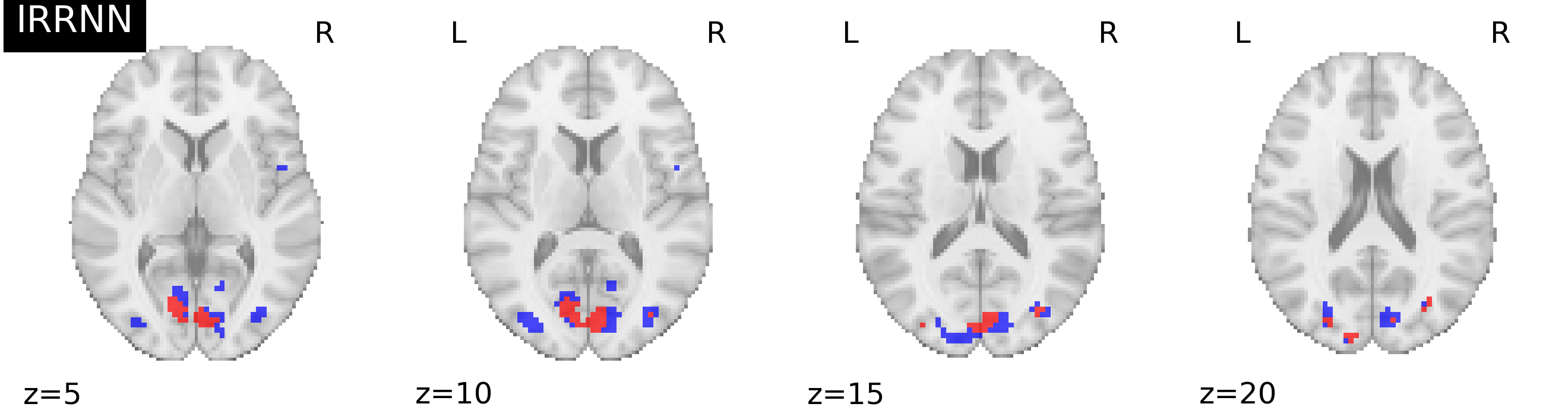}
\caption{\label{fig:brain-abide} ABIDE.}
\end{subfigure}
\hfill
\begin{subfigure}{0.49\textwidth}
\centering
        \includegraphics[width=\textwidth]{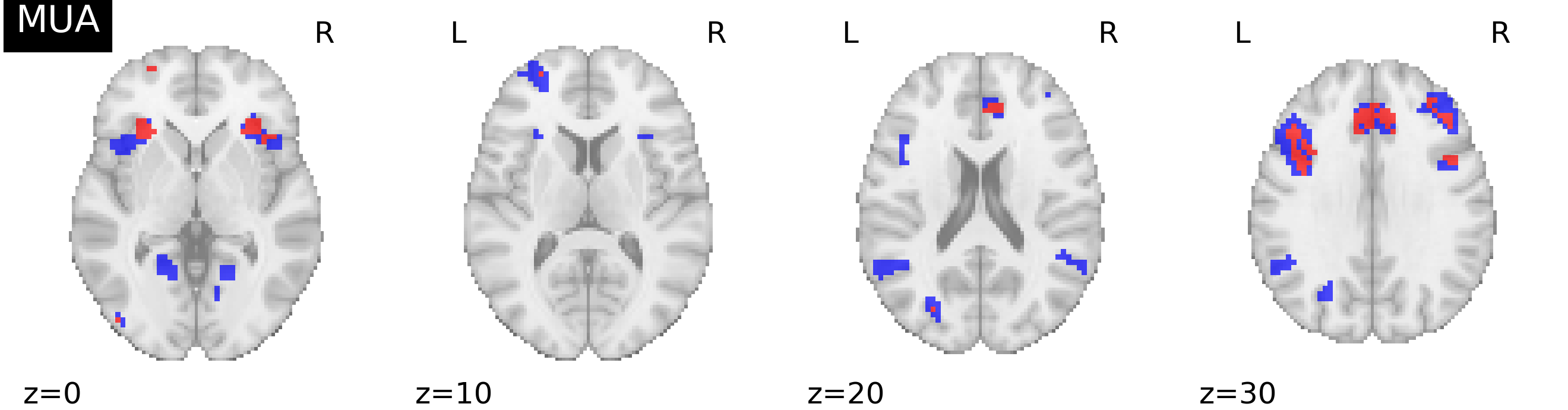}
        \includegraphics[width=\textwidth]{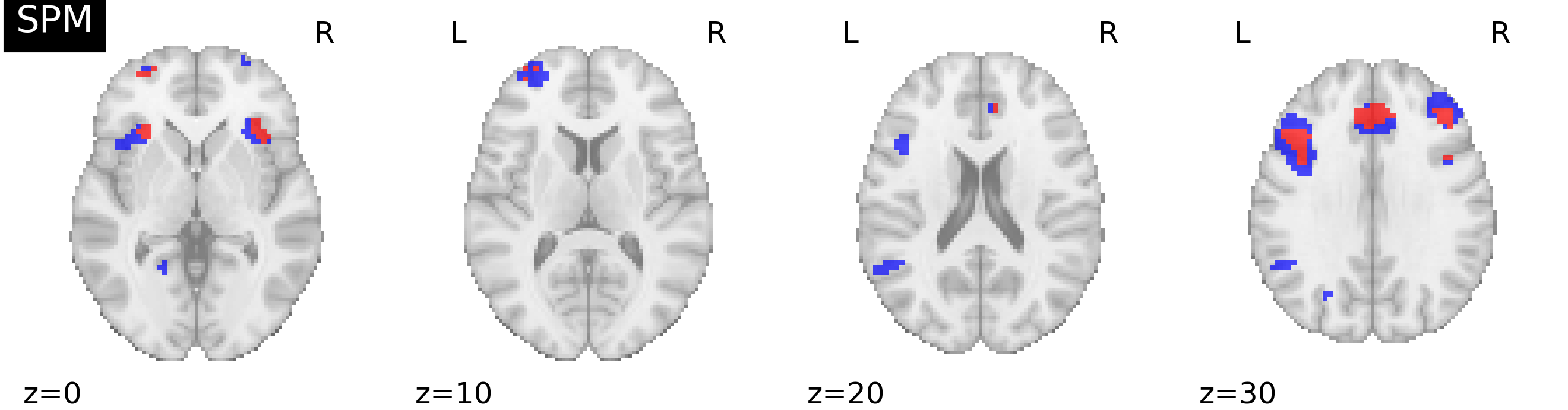}
        \includegraphics[width=\textwidth]{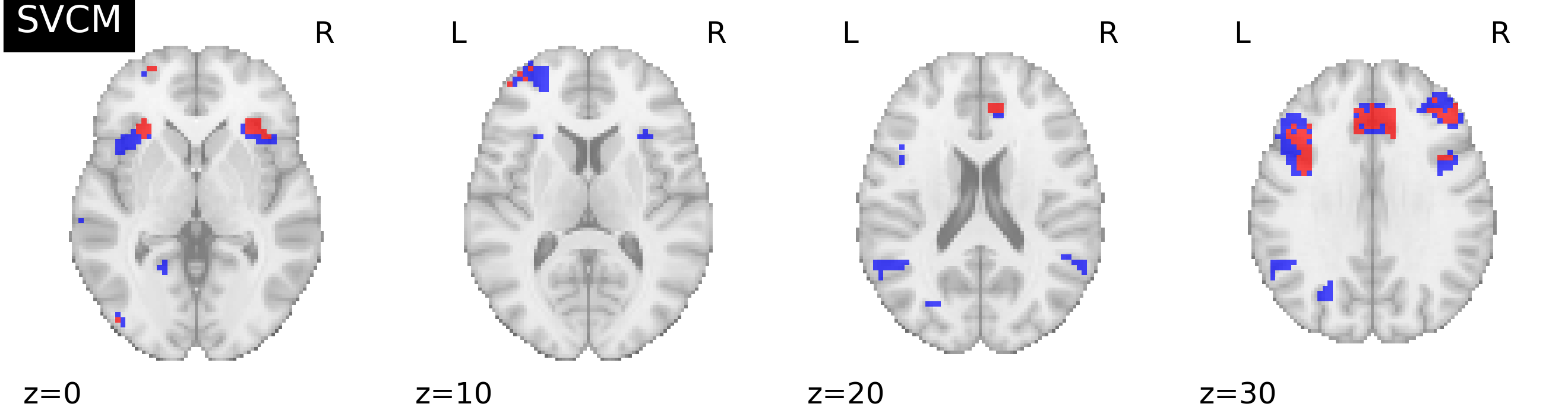}
        \includegraphics[width=\textwidth]{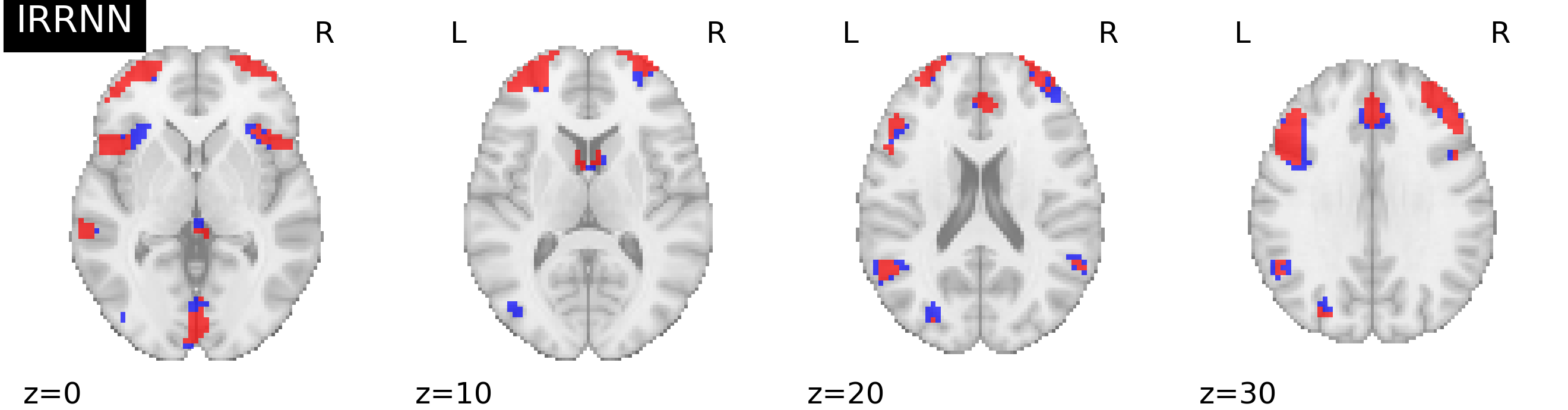}
\caption{\label{fig:brain-abcd} ABCD}
\end{subfigure}
\caption{
  Voxels by various methods in the brain fMRI studies.
  Red voxels are those selected in the all-site analysis that are reproducible in the single-site analyses,
  while blue voxels are those selected in the all-site analysis that are not reproducible in the single-site analyses.
}
\end{figure}

We next evaluate the selection accuracy. Again, since the true parameter functions are unknown for the real data, we evaluate the methods by their reproducibility in selecting important voxels. That is, we first obtain the set of selected voxels from the analysis of the data combined from all imaging sites. We then identify the set of selected voxels from the analysis of each individual imaging site. We claim a selected voxel from the all-site analysis is reproducible if it is selected in at least five single-site analyses. \Cref{tbl:sitewise-sel} reports the proportion of the reproducible voxels by all the methods. Our method achieves the highest proportion of reproducible voxels, which doubles that of SVCM, while the reproducible proportion for MUA and SPM are almost zero. The selection reproducibility is visualized in \Cref{fig:brain-abide}.

\FloatBarrier
\begin{table}[t!]
\caption{\label{tbl:sitewise-sel}
  Selection consistency in the brain fMRI studies,
  as measured by the proportion of voxels selected in the all-site analysis that are also consistently selected in single-site analyses.}
\centering
\begin{tabular}{lrrrr} \toprule
Study & MUA   & SPM   & SVCM  & IRRNN \\ \midrule
ABIDE & 0.021 & 0.000 & 0.168 & 0.365 \\
ABCD  & 0.397 & 0.423 & 0.428 & 0.655 \\ \bottomrule
\end{tabular}
\end{table}

We also present a summary of the selected voxels from all imaging sites. We first map those voxels to a set of predefined brain anatomical regions following the widely used automated anatomical labeling (AAL) atlas \citep{tzourio2002automated}. Then we further group these regions into a set of functional networks \citep{power2011functional}. \Cref{tbl:region} reports the top selected regions and functional networks. These findings are consistent with the literature on cognitive ability. In particular, we have selected a high proportion of voxels in the regions of occipital lobe \citep{goriounova2019genes, yoon2017brain, simard2015autistic, menary2013associations}, the calcarine fissure and surrounding cortex \citep{yu2008white}, and the cuneus \citep{schnack2015changes, haier2004structural, song2008brain}. In addition, these regions all belong to the visual functional network, which has been found to be closely related to cognitive ability \citep{dubois2018distributed, hearne2016functional}. Compared to the alternative methods, IRRNN's reproducibility is higher in most of the top regions and functional networks. This result shows the selection consistency of IRRNN inside biologically meaningful imaging regions.

\subsection{ABCD study}

The ABCD study aims at delineating the associations between cognitive behaviors and brain development \citep{casey2018adolescent}. In our analysis, we use the annual release 1.1 that consists of minimally preprocessed two-back task-based fMRI data from 20 imaging sites \citep{hagler2019image}. The response image we analyze is the contrast map of the two-back task, which has been consistently found to engage brain regions for memory regulation processes and cognitive functions \citep{barch2013function}. The imaging dimension is $61 \times 73 \times 61$. The $J=4$ covariates include the general cognitive ability component score \citep{sripada2019prediction}, the psychiatric diagnostic score, age, and sex. The number of subjects is $N = 1991$. 

Similar to the analysis of the ABIDE data, we apply IRRNN and compare it with MUA, SPM, and SVCM. We evaluate the estimation accuracy, by reporting the cross-validation of the mean squared error of the recovered response image. As shown in \Cref{tab:sitewise-fit}, our method is the most accurate in almost all the experiments. In the most extreme case, our method reduces the MSE of the alternative methods by more than 16-fold. Even in the cases where IRRNN is not the best performer, we still produces estimates with comparable accuracy. Moreover, we evaluate the selection accuracy by reporting the proportion of the reproducible voxel, as listed in \Cref{tbl:sitewise-sel} and visualized in \Cref{fig:brain-abcd}. Our method again achieves the highest proportion of reproducible voxels. \Cref{tbl:region} reports the selected top regions and functional networks. In particular, we select a high proportion of voxels in the regions known to be associated with CA, including the parietal lobe \citep{haier2005neuroanatomy, woolgar2010fluid, jung2007parieto}, the frontal lobe \citep{duncan1996intelligence, roca2010executive}, and the precuneus \citep{basten2015smart, jauk2015gray}. Moreover, the memory retrieval functional network has the highest selection proportion, which is consistent with the fact that the ABCD response images used in our analysis
are task-based contrast maps for activities designed to engage working memory. The selection proportion for the dorsal attention functional network is also high, which has been found in the literature to be associated with both general CA \citep{hilger2020temporal} and working memory capacity \citep{majerus2018dorsal, broadway2010validating, gray2017structure}.
Other top functional networks include frontal-parietal task control \citep{uddin2019towards, zanto2013fronto} and salience \citep{hilger2017efficient, liang2016topologically}, which is also consistent with the findings in existing works regarding their associations with CA and working memory. Finally, the reproducibility rate of IRRNN is much higher than 
that of the other methods in most of the AAL regions and networks, making IRRNN the most stable and consistent method for selection within biologically defined brain regions.

\begin{table}[t!]
\centering
\footnotesize{
\caption{\label{tbl:region} 
    Voxel selection and reproducibility within AAL regions and functional networks
    in the brain fMRI data.
    Inside each row for each method,
    the first column is the name of the region/network,
    the second column shows the proportion of voxels inside the region/network
    that are selected in the all-site analysis,
    and the third column (in parentheses) reports the proportion of the voxels
    in the all-site analysis that are reproducible in the single-site analyses.}
\footnotesize
\setlength{\tabcolsep}{2.0pt}
\renewcommand{\arraystretch}{0.5}
\begin{tabular}{lrrlrrlrrlrr}
    \toprule
    \multicolumn{3}{c}{MUA} & \multicolumn{3}{c}{SPM} & \multicolumn{3}{c}{SVCM} & \multicolumn{3}{c}{IRRNN} \\
    \midrule
    \multicolumn{12}{c}{AAL regions in ABIDE} \\
    \midrule
    Occ.Mid.R        & 11 & (10) & Occ.Mid.L       & 13 & (00) & Rec.R            & 13 & (00) & Cal.R           & 17 & (40) \\
    Cal.R            & 10 & (00) & Cal.R           & 12 & (00) & Rec.L            & 12 & (25) & Cal.L           & 15 & (67) \\
    Rec.L            & 09 & (00) & Occ.Sup.L       & 10 & (00) & Occ.Mid.R        & 10 & (23) & Cun.R           & 12 & (27) \\
    Occ.Mid.L        & 07 & (00) & Occ.Mid.R       & 10 & (00) & Occ.Mid.L        & 09 & (00) & Occ.Mid.R       & 12 & (46) \\
    Occ.Sup.L        & 06 & (04) & Cun.R           & 08 & (00) & Occ.Sup.L        & 08 & (04) & Occ.Mid.L       & 07 & (14) \\
    Sup.Mot.Are.R    & 06 & (00) & Cal.L           & 05 & (00) & Cal.R            & 06 & (03) & Occ.Sup.L       & 06 & (39) \\
    Cal.L            & 05 & (00) & Sup.Mot.Are.R   & 05 & (00) & Fro.Med.Orb.L    & 05 & (38) & Cun.L           & 05 & (35) \\
    Rec.R            & 04 & (00) & Cun.L           & 04 & (00) & Fro.Sup.Orb.R    & 05 & (83) & Fus.L           & 03 & (06) \\
    Cun.L            & 04 & (00) & Rec.L           & 03 & (00) & Tem.Pol.Mid.L    & 04 & (00) & Lin.L           & 03 & (21) \\
    Occ.Inf.R        & 04 & (00) & Rec.R           & 02 & (00) & Sup.Mot.Are.R    & 04 & (00) & Sup.Mot.Are.R   & 02 & (22) \\
    \midrule
    \multicolumn{12}{c}{Functional networks in ABIDE} \\
    \midrule
             Vis     & 04 & (03) &             Vis & 06 & (00) &              Vis & 04 & (07) &             Vis & 07 & (38) \\
             Ven.Att & 01 & (00) &         Ven.Att & 01 & (00) &      Sen.Som.Han & 01 & (24) &         Def.Mod & 01 & (46) \\
             Tas.Con & 01 & (00) & Cin.Ope.Tas.Con & 01 & (00) &  Cin.Ope.Tas.Con & 01 & (04) &         Ven.Att & 00 & (22) \\
             Som.Han & 01 & (02) &         Def.Mod & 00 & (00) &          Ven.Att & 01 & (00) & Cin.Ope.Tas.Con & 00 & (20) \\
             Def.Mod & 01 & (07) &     Sen.Som.Han & 00 & (00) &          Def.Mod & 01 & (21) &     Sen.Som.Han & 00 & (22) \\
    \midrule
    \multicolumn{12}{c}{AAL regions in ABCD} \\
    \midrule
           Par.Inf.L & 49 & (42) &       Par.Inf.L & 55 & (40) &        Par.Inf.L & 51 & (42) &       Par.Inf.L & 51 & (74) \\
           Par.Sup.L & 40 & (23) &       Par.Sup.L & 50 & (30) &        Par.Sup.R & 47 & (46) &   Fro.Mid.Orb.L & 44 & (00) \\
               Pre.L & 35 & (59) &       Par.Sup.R & 47 & (35) &        Par.Sup.L & 46 & (33) &       Par.Sup.L & 42 & (70) \\
           Par.Inf.R & 34 & (56) &           Pre.L & 40 & (61) &        Par.Inf.R & 39 & (56) &       Fro.Mid.L & 36 & (79) \\
           Inf.Ope.L & 33 & (40) &       Par.Inf.R & 38 & (54) &            Pre.L & 35 & (63) &       Par.Sup.R & 36 & (25) \\
           Par.Sup.R & 32 & (13) &   Fro.Inf.Ope.L & 33 & (37) &            Pre.R & 32 & (71) &   Fro.Inf.Ope.L & 33 & (73) \\
               Pre.R & 31 & (65) &           Pre.R & 33 & (68) &    Fro.Inf.Ope.L & 31 & (31) &   Fro.Sup.Orb.L & 32 & (97) \\
               Pre.L & 29 & (59) &       Fro.Mid.L & 28 & (49) &        Fro.Mid.L & 29 & (53) &           Pre.L & 31 & (83) \\
               Lin.L & 25 & (12) &           Pre.L & 27 & (53) &            Pre.L & 28 & (57) &   Fro.Mid.Orb.R & 31 & (96) \\
           Fro.Mid.L & 24 & (42) &   Sup.Mot.Are.L & 24 & (82) &            Lin.L & 23 & (03) &       Fro.Mid.R & 29 & (75) \\
    \midrule
    \multicolumn{12}{c}{Functional networks in ABCD} \\
    \midrule
             Mem.Ret & 35 & (59) &         Mem.Ret & 40 & (61) &          Mem.Ret & 35 & (63) &         Mem.Ret & 31 & (83) \\
             Dor.Att & 22 & (43) &         Dor.Att & 25 & (49) &          Dor.Att & 23 & (49) & Fro.Par.Tas.Con & 24 & (75) \\
     Fro.Par.Tas.Con & 19 & (43) & Fro.Par.Tas.Con & 19 & (47) &  Fro.Par.Tas.Con & 20 & (48) &             Sal & 23 & (75) \\
                 Sal & 18 & (42) &             Sal & 17 & (47) &              Sal & 19 & (49) &         Dor.Att & 21 & (77) \\
     Cin.Ope.Tas.Con & 17 & (53) & Cin.Ope.Tas.Con & 16 & (62) &  Cin.Ope.Tas.Con & 16 & (53) &         Def.Mod & 14 & (72) \\
    \bottomrule
\end{tabular}
}
\end{table}

%% file: discussion.tex
\section{Discussion} 
\label{sec:discussion}

In this work,
we have presented a novel image-on-scalar regression model
based on deep neural networks.
From the perspective of functional data analysis,
our model uses multi-layer feed-forward neural networks
to approximate the spatially varying coefficient functions
of the main effects, individual deviations, and noise variance.
Although conceptually straightforward,
our model is capable of adapting to
a wide variety of spatial correlation patterns,
including not only smooth transitions
but also jump discontinuities across
the spatial volume.
We have provided an algorithm for model fitting and selection
that takes advantage of the high-dimensionality
of the imaging data.
This estimation procedure
has been proved to possess theoretically guaranteed
convergence properties.

In our theoretical analysis,
we have derived $L_2$ estimation error bounds
for the main effects, individual deviations, and noise variance.
Based on existing works
on nonparametric regression with deep neural networks,
our theoretical results extend them from globally smooth functions
to piecewise smooth functions.
Moreover, for our model selection procedure,
we have demonstrated their selection consistency
and proved $L_0$ sign error bounds.
In addition, we have shown the best neural network complexity,
as a function of the number of images and voxels,
that gives the fastest convergence rate.

In our extensive simulation studies,
we have designed complex spatial images
to test IRRNN against multiple existing 
image response regression methods.
IRRNN has successfully eliminated
a great proportion of the noise
and learned most of the underlying heterogeneous patterns
in the main effects.
Moreover, IRRNN has been shown to be effective
in exploiting the increasing imaging dimensions,
compared to existing methods,
both in terms of estimation accuracy
and selection accuracy.
For the analysis of brain fMRI data,
IRRNN has produced more accurate estimates
in the cross-validations across experimental sites.
The advantage of IRRNN over the baseline methods
is more prominent
when the estimation task is more difficult.
In addition,
our model has achieved the highest
selection reproducibility,
as measured by the proportion of voxels that are
consistently selected in the single-site analyses.
Finally,
as we break down the results into AAL regions
and functional networks,
the top regions and networks selected by IRRNN
are supported by findings in existing works.
The reproducibility rate in each region or network
is also much higher for IRRNN than the alternative methods.

The method proposed in this article serves as an example on
the integration of modern deep learning tools
with classical statistical frameworks,
such that data-driven, highly flexible modeling characteristics
can be enjoyed along with theoretical guarantees.
We believe this strategy represents an important direction
for the analysis of next-generation high-dimensional complex data.
For future works,
we envision multiple extensions of our proposed approach.
For example,
in our current setting,
although the imaging resolution is high,
the number of covariates is fixed at a constant value.
We could extend the current theoretical results
and modify the algorithm
to accommodate an increasing number of covariates,
which is common in applications such as imaging genetics.
In addition, the images are currently assumed to be single-channel.
This condition can be potentially generalized
to multi-channel images,
such as those in spatial transcriptomic data.

%% file: main.bbl
\begin{thebibliography}{}

\bibitem[Barch et~al., 2013]{barch2013function}
Barch, D.~M., Burgess, G.~C., Harms, M.~P., Petersen, S.~E., Schlaggar, B.~L.,
  Corbetta, M., Glasser, M.~F., Curtiss, S., Dixit, S., Feldt, C., et~al.
  (2013).
\newblock Function in the human connectome: task-fmri and individual
  differences in behavior.
\newblock {\em Neuroimage}, 80:169--189.

\bibitem[Barron, 1994]{barron1994approximation}
Barron, A.~R. (1994).
\newblock Approximation and estimation bounds for artificial neural networks.
\newblock {\em Machine Learning}, 14(1):115--133.

\bibitem[Basten et~al., 2015]{basten2015smart}
Basten, U., Hilger, K., and Fiebach, C.~J. (2015).
\newblock Where smart brains are different: A quantitative meta-analysis of
  functional and structural brain imaging studies on intelligence.
\newblock {\em Intelligence}, 51:10--27.

\bibitem[Bauer et~al., 2019]{bauer2019deep}
Bauer, B., Kohler, M., et~al. (2019).
\newblock On deep learning as a remedy for the curse of dimensionality in
  nonparametric regression.
\newblock {\em Annals of Statistics}, 47(4):2261--2285.

\bibitem[Bottou, 2010]{bottou2010large}
Bottou, L. (2010).
\newblock Large-scale machine learning with stochastic gradient descent.
\newblock In {\em Proceedings of COMPSTAT'2010}, pages 177--186. Springer.

\bibitem[Broadway and Engle, 2010]{broadway2010validating}
Broadway, J.~M. and Engle, R.~W. (2010).
\newblock Validating running memory span: Measurement of working memory
  capacity and links with fluid intelligence.
\newblock {\em Behavior Research Methods}, 42(2):563--570.

\bibitem[Bussas et~al., 2017]{bussas2017varying}
Bussas, M., Sawade, C., K{\"u}hn, N., Scheffer, T., and Landwehr, N. (2017).
\newblock Varying-coefficient models for geospatial transfer learning.
\newblock {\em Machine Learning}, 106(9-10):1419--1440.

\bibitem[Casey et~al., 2018]{casey2018adolescent}
Casey, B., Cannonier, T., Conley, M.~I., Cohen, A.~O., Barch, D.~M., Heitzeg,
  M.~M., Soules, M.~E., Teslovich, T., Dellarco, D.~V., Garavan, H., et~al.
  (2018).
\newblock The adolescent brain cognitive development (abcd) study: imaging
  acquisition across 21 sites.
\newblock {\em Developmental Cognitive Neuroscience}, 32:43--54.

\bibitem[Chen et~al., 2019]{chen2019non}
Chen, H., Raskutti, G., and Yuan, M. (2019).
\newblock Non-convex projected gradient descent for generalized low-rank tensor
  regression.
\newblock {\em The Journal of Machine Learning Research}, 20(1):172--208.

\bibitem[Chen et~al., 2020]{chen2020nonlinear}
Chen, Y., Gao, Q., Liang, F., and Wang, X. (2020).
\newblock Nonlinear variable selection via deep neural networks.
\newblock {\em Journal of Computational and Graphical Statistics}, pages 1--9.

\bibitem[Chen et~al., 2016]{chen2016local}
Chen, Y., Wang, X., Kong, L., and Zhu, H. (2016).
\newblock Local region sparse learning for image-on-scalar regression.
\newblock {\em arXiv preprint arXiv:1605.08501}.

\bibitem[Chumbley and Friston, 2009]{chumbley2009false}
Chumbley, J.~R. and Friston, K.~J. (2009).
\newblock False discovery rate revisited: Fdr and topological inference using
  gaussian random fields.
\newblock {\em Neuroimage}, 44(1):62--70.

\bibitem[Craddock et~al., 2013]{cpac2013}
Craddock, C., Sikka, S., Cheung, B., Khanuja, R., Ghosh, S.~S., Yan, C., Li,
  Q., Lurie, D., Vogelstein, J., Burns, R., Colcombe, S., Mennes, M., Kelly,
  C., Di~Martino, A., Castellanos, F.~X., and Milham, M. (2013).
\newblock Towards automated analysis of connectomes: The configurable pipeline
  for the analysis of connectomes (c-pac).
\newblock {\em Frontiers in Neuroinformatics}, (42).

\bibitem[Di~Martino et~al., 2014]{di2014autism}
Di~Martino, A., Yan, C.-G., Li, Q., Denio, E., Castellanos, F.~X., Alaerts, K.,
  Anderson, J.~S., Assaf, M., Bookheimer, S.~Y., Dapretto, M., et~al. (2014).
\newblock The autism brain imaging data exchange: towards a large-scale
  evaluation of the intrinsic brain architecture in autism.
\newblock {\em Molecular Psychiatry}, 19(6):659--667.

\bibitem[Dubois et~al., 2018]{dubois2018distributed}
Dubois, J., Galdi, P., Paul, L.~K., and Adolphs, R. (2018).
\newblock A distributed brain network predicts general intelligence from
  resting-state human neuroimaging data.
\newblock {\em Philosophical Transactions of the Royal Society B: Biological
  Sciences}, 373(1756):20170284.

\bibitem[Duncan et~al., 1996]{duncan1996intelligence}
Duncan, J., Emslie, H., Williams, P., Johnson, R., and Freer, C. (1996).
\newblock Intelligence and the frontal lobe: The organization of goal-directed
  behavior.
\newblock {\em Cognitive psychology}, 30(3):257--303.

\bibitem[Eldan and Shamir, 2016]{eldan2016power}
Eldan, R. and Shamir, O. (2016).
\newblock The power of depth for feedforward neural networks.
\newblock In {\em Conference on learning theory}, pages 907--940. PMLR.

\bibitem[Fan et~al., 2019]{fan2019selective}
Fan, J., Ma, C., and Zhong, Y. (2019).
\newblock A selective overview of deep learning.
\newblock {\em arXiv preprint arXiv:1904.05526}.

\bibitem[Feng and Simon, 2017]{feng2017sparse}
Feng, J. and Simon, N. (2017).
\newblock Sparse-input neural networks for high-dimensional nonparametric
  regression and classification.
\newblock {\em arXiv preprint arXiv:1711.07592}.

\bibitem[Friston, 2003]{friston2003statistical}
Friston, K.~J. (2003).
\newblock Statistical parametric mapping.
\newblock In {\em Neuroscience databases}, pages 237--250. Springer.

\bibitem[Goodfellow et~al., 2016]{goodfellow2016deep}
Goodfellow, I., Bengio, Y., Courville, A., and Bengio, Y. (2016).
\newblock {\em Deep learning}.
\newblock MIT press Cambridge.

\bibitem[Goriounova and Mansvelder, 2019]{goriounova2019genes}
Goriounova, N.~A. and Mansvelder, H.~D. (2019).
\newblock Genes, cells and brain areas of intelligence.
\newblock {\em Frontiers in human neuroscience}, 13:44.

\bibitem[Gray et~al., 2017]{gray2017structure}
Gray, S., Green, S., Alt, M., Hogan, T., Kuo, T., Brinkley, S., and Cowan, N.
  (2017).
\newblock The structure of working memory in young children and its relation to
  intelligence.
\newblock {\em Journal of Memory and Language}, 92:183--201.

\bibitem[Hagler~Jr et~al., 2019]{hagler2019image}
Hagler~Jr, D.~J., Hatton, S., Cornejo, M.~D., Makowski, C., Fair, D.~A., Dick,
  A.~S., Sutherland, M.~T., Casey, B., Barch, D.~M., Harms, M.~P., et~al.
  (2019).
\newblock Image processing and analysis methods for the adolescent brain
  cognitive development study.
\newblock {\em NeuroImage}, 202:116091.

\bibitem[Haier et~al., 2004]{haier2004structural}
Haier, R.~J., Jung, R.~E., Yeo, R.~A., Head, K., and Alkire, M.~T. (2004).
\newblock Structural brain variation and general intelligence.
\newblock {\em Neuroimage}, 23(1):425--433.

\bibitem[Haier et~al., 2005]{haier2005neuroanatomy}
Haier, R.~J., Jung, R.~E., Yeo, R.~A., Head, K., and Alkire, M.~T. (2005).
\newblock The neuroanatomy of general intelligence: sex matters.
\newblock {\em NeuroImage}, 25(1):320--327.

\bibitem[He et~al., 2019]{he2019selective}
He, K., Xu, H., and Kang, J. (2019).
\newblock A selective overview of feature screening methods with applications
  to neuroimaging data.
\newblock {\em Wiley Interdisciplinary Reviews: Computational Statistics},
  11(2):e1454.

\bibitem[Hearne et~al., 2016]{hearne2016functional}
Hearne, L.~J., Mattingley, J.~B., and Cocchi, L. (2016).
\newblock Functional brain networks related to individual differences in human
  intelligence at rest.
\newblock {\em Scientific reports}, 6(1):1--8.

\bibitem[Hilger et~al., 2017]{hilger2017efficient}
Hilger, K., Ekman, M., Fiebach, C.~J., and Basten, U. (2017).
\newblock Efficient hubs in the intelligent brain: Nodal efficiency of hub
  regions in the salience network is associated with general intelligence.
\newblock {\em Intelligence}, 60:10--25.

\bibitem[Hilger et~al., 2020]{hilger2020temporal}
Hilger, K., Fukushima, M., Sporns, O., and Fiebach, C.~J. (2020).
\newblock Temporal stability of functional brain modules associated with human
  intelligence.
\newblock {\em Human brain mapping}, 41(2):362--372.

\bibitem[Jauk et~al., 2015]{jauk2015gray}
Jauk, E., Neubauer, A.~C., Dunst, B., Fink, A., and Benedek, M. (2015).
\newblock Gray matter correlates of creative potential: A latent variable
  voxel-based morphometry study.
\newblock {\em NeuroImage}, 111:312--320.

\bibitem[Jung and Haier, 2007]{jung2007parieto}
Jung, R.~E. and Haier, R.~J. (2007).
\newblock The parieto-frontal integration theory (p-fit) of intelligence:
  converging neuroimaging evidence.
\newblock {\em Behavioral and Brain Sciences}, 30(2):135.

\bibitem[Kohler and Krzy{\.z}ak, 2017]{kohler2017nonparametric}
Kohler, M. and Krzy{\.z}ak, A. (2017).
\newblock Nonparametric regression based on hierarchical interaction models.
\newblock {\em IEEE Transactions on Information Theory}, 63(3):1620--1630.

\bibitem[LeCun et~al., 2015]{lecun2015deep}
LeCun, Y., Bengio, Y., and Hinton, G. (2015).
\newblock Deep learning.
\newblock {\em nature}, 521(7553):436--444.

\bibitem[Li et~al., 2017]{LiZhu2017}
Li, J., Huang, C., Hongtu, Z., and for~the Alzheimer’s Disease
  Neuroimaging~Initiative (2017).
\newblock A functional varying-coefficient single-index model for functional
  response data.
\newblock {\em Journal of the American Statistical Association},
  112(519):1169--1181.

\bibitem[Li and Zhang, 2017]{li2017parsimonious}
Li, L. and Zhang, X. (2017).
\newblock Parsimonious tensor response regression.
\newblock {\em Journal of the American Statistical Association},
  112(519):1131--1146.

\bibitem[Li et~al., 2020]{li2020sparse}
Li, X., Wang, L., Wang, H.~J., and Initiative, A. D.~N. (2020).
\newblock Sparse learning and structure identification for
  ultrahigh-dimensional image-on-scalar regression.
\newblock {\em Journal of the American Statistical Association}, pages 1--15.

\bibitem[Liang et~al., 2016]{liang2016topologically}
Liang, X., Zou, Q., He, Y., and Yang, Y. (2016).
\newblock Topologically reorganized connectivity architecture of default-mode,
  executive-control, and salience networks across working memory task loads.
\newblock {\em Cerebral cortex}, 26(4):1501--1511.

\bibitem[Majerus et~al., 2018]{majerus2018dorsal}
Majerus, S., P{\'e}ters, F., Bouffier, M., Cowan, N., and Phillips, C. (2018).
\newblock The dorsal attention network reflects both encoding load and
  top--down control during working memory.
\newblock {\em Journal of Cognitive Neuroscience}, 30(2):144--159.

\bibitem[McCaffrey and Gallant, 1994]{mccaffrey1994convergence}
McCaffrey, D.~F. and Gallant, A.~R. (1994).
\newblock Convergence rates for single hidden layer feedforward networks.
\newblock {\em Neural Networks}, 7(1):147--158.

\bibitem[Menary et~al., 2013]{menary2013associations}
Menary, K., Collins, P.~F., Porter, J.~N., Muetzel, R., Olson, E.~A., Kumar,
  V., Steinbach, M., Lim, K.~O., and Luciana, M. (2013).
\newblock Associations between cortical thickness and general intelligence in
  children, adolescents and young adults.
\newblock {\em Intelligence}, 41(5):597--606.

\bibitem[Power et~al., 2011]{power2011functional}
Power, J.~D., Cohen, A.~L., Nelson, S.~M., Wig, G.~S., Barnes, K.~A., Church,
  J.~A., Vogel, A.~C., Laumann, T.~O., Miezin, F.~M., Schlaggar, B.~L., et~al.
  (2011).
\newblock Functional network organization of the human brain.
\newblock {\em Neuron}, 72(4):665--678.

\bibitem[Qiu, 2007]{qiu2007jump}
Qiu, P. (2007).
\newblock Jump surface estimation, edge detection, and image restoration.
\newblock {\em Journal of the American Statistical Association},
  102(478):745--756.

\bibitem[Rabusseau and Kadri, 2016]{rabusseau2016}
Rabusseau, G. and Kadri, H. (2016).
\newblock Low-rank regression with tensor responses.
\newblock In {\em Advances in Neural Information Processing Systems}.

\bibitem[Raskutti et~al., 2019]{raskutti2019convex}
Raskutti, G., Yuan, M., Chen, H., et~al. (2019).
\newblock Convex regularization for high-dimensional multiresponse tensor
  regression.
\newblock {\em The Annals of Statistics}, 47(3):1554--1584.

\bibitem[Roca et~al., 2010]{roca2010executive}
Roca, M., Parr, A., Thompson, R., Woolgar, A., Torralva, T., Antoun, N., Manes,
  F., and Duncan, J. (2010).
\newblock Executive function and fluid intelligence after frontal lobe lesions.
\newblock {\em Brain}, 133(1):234--247.

\bibitem[Schmidt-Hieber et~al., 2020]{schmidt2020nonparametric}
Schmidt-Hieber, J. et~al. (2020).
\newblock Nonparametric regression using deep neural networks with relu
  activation function.
\newblock {\em Annals of Statistics}, 48(4):1875--1897.

\bibitem[Schnack et~al., 2015]{schnack2015changes}
Schnack, H.~G., Van~Haren, N.~E., Brouwer, R.~M., Evans, A., Durston, S.,
  Boomsma, D.~I., Kahn, R.~S., and Hulshoff~Pol, H.~E. (2015).
\newblock Changes in thickness and surface area of the human cortex and their
  relationship with intelligence.
\newblock {\em Cerebral cortex}, 25(6):1608--1617.

\bibitem[Shi and Kang, 2015]{shi2015thresholded}
Shi, R. and Kang, J. (2015).
\newblock Thresholded multiscale gaussian processes with application to
  bayesian feature selection for massive neuroimaging data.
\newblock {\em arXiv preprint arXiv:1504.06074}.

\bibitem[Simard et~al., 2015]{simard2015autistic}
Simard, I., Luck, D., Mottron, L., Zeffiro, T.~A., and Souli{\`e}res, I.
  (2015).
\newblock Autistic fluid intelligence: Increased reliance on visual functional
  connectivity with diminished modulation of coupling by task difficulty.
\newblock {\em NeuroImage: Clinical}, 9:467--478.

\bibitem[Song et~al., 2008]{song2008brain}
Song, M., Zhou, Y., Li, J., Liu, Y., Tian, L., Yu, C., and Jiang, T. (2008).
\newblock Brain spontaneous functional connectivity and intelligence.
\newblock {\em Neuroimage}, 41(3):1168--1176.

\bibitem[Sripada et~al., 2019]{sripada2019prediction}
Sripada, C., Rutherford, S., Angstadt, M., Thompson, W.~K., Luciana, M.,
  Weigard, A., Hyde, L.~H., and Heitzeg, M. (2019).
\newblock Prediction of neurocognition in youth from resting state fmri.
\newblock {\em Molecular Psychiatry}, pages 1--9.

\bibitem[Stone, 1982]{stone1982optimal}
Stone, C.~J. (1982).
\newblock Optimal global rates of convergence for nonparametric regression.
\newblock {\em The annals of statistics}, pages 1040--1053.

\bibitem[Sun and Li, 2017]{sun2017store}
Sun, W.~W. and Li, L. (2017).
\newblock Store: sparse tensor response regression and neuroimaging analysis.
\newblock {\em The Journal of Machine Learning Research}, 18(1):4908--4944.

\bibitem[Telgarsky, 2016]{telgarsky2016benefits}
Telgarsky, M. (2016).
\newblock Benefits of depth in neural networks.
\newblock In {\em Conference on learning theory}, pages 1517--1539. PMLR.

\bibitem[Tzourio-Mazoyer et~al., 2002]{tzourio2002automated}
Tzourio-Mazoyer, N., Landeau, B., Papathanassiou, D., Crivello, F., Etard, O.,
  Delcroix, N., Mazoyer, B., and Joliot, M. (2002).
\newblock Automated anatomical labeling of activations in spm using a
  macroscopic anatomical parcellation of the mni mri single-subject brain.
\newblock {\em Neuroimage}, 15(1):273--289.

\bibitem[Uddin et~al., 2019]{uddin2019towards}
Uddin, L.~Q., Yeo, B.~T., and Spreng, R.~N. (2019).
\newblock Towards a universal taxonomy of macro-scale functional human brain
  networks.
\newblock {\em Brain topography}, 32(6):926--942.

\bibitem[Woolgar et~al., 2010]{woolgar2010fluid}
Woolgar, A., Parr, A., Cusack, R., Thompson, R., Nimmo-Smith, I., Torralva, T.,
  Roca, M., Antoun, N., Manes, F., and Duncan, J. (2010).
\newblock Fluid intelligence loss linked to restricted regions of damage within
  frontal and parietal cortex.
\newblock {\em Proceedings of the National Academy of Sciences},
  107(33):14899--14902.

\bibitem[Yoon et~al., 2017]{yoon2017brain}
Yoon, Y.~B., Shin, W.-G., Lee, T.~Y., Hur, J.-W., Cho, K. I.~K., Sohn, W.~S.,
  Kim, S.-G., Lee, K.-H., and Kwon, J.~S. (2017).
\newblock Brain structural networks associated with intelligence and visuomotor
  ability.
\newblock {\em Scientific reports}, 7(1):1--9.

\bibitem[Yu et~al., 2008]{yu2008white}
Yu, C., Li, J., Liu, Y., Qin, W., Li, Y., Shu, N., Jiang, T., and Li, K.
  (2008).
\newblock White matter tract integrity and intelligence in patients with mental
  retardation and healthy adults.
\newblock {\em Neuroimage}, 40(4):1533--1541.

\bibitem[Yu et~al., 2020]{yu2020multivariate}
Yu, S., Wang, G., Wang, L., and Yang, L. (2020).
\newblock Multivariate spline estimation and inference for image-on-scalar
  regression.
\newblock {\em Statistica Sinica}.

\bibitem[Yue et~al., 2010]{yue2010adaptive}
Yue, Y., Loh, J.~M., and Lindquist, M.~A. (2010).
\newblock Adaptive spatial smoothing of fmri images.
\newblock {\em Statistics and its Interface}, 3:3--13.

\bibitem[Zanto and Gazzaley, 2013]{zanto2013fronto}
Zanto, T.~P. and Gazzaley, A. (2013).
\newblock Fronto-parietal network: flexible hub of cognitive control.
\newblock {\em Trends in cognitive sciences}, 17(12):602--603.

\bibitem[Zhao and Yu, 2006]{zhao2006model}
Zhao, P. and Yu, B. (2006).
\newblock On model selection consistency of lasso.
\newblock {\em The Journal of Machine Learning Research}, 7:2541--2563.

\bibitem[Zhou, 2010]{zhou2010thresholded}
Zhou, S. (2010).
\newblock Thresholded lasso for high dimensional variable selection and
  statistical estimation.
\newblock {\em arXiv preprint arXiv:1002.1583}.

\bibitem[Zhu et~al., 2014]{zhu2014spatially}
Zhu, H., Fan, J., and Kong, L. (2014).
\newblock Spatially varying coefficient model for neuroimaging data with jump
  discontinuities.
\newblock {\em Journal of the American Statistical Association},
  109(507):1084--1098.

\end{thebibliography}
